\documentclass[journal]{IEEEtran}
\usepackage{ifpdf}

\usepackage{cite}

\usepackage{float}
\usepackage{graphicx}
\usepackage{caption}
\usepackage[skip=0cm,list=true,labelfont=it]{subcaption}
\usepackage{xspace}
\usepackage{multirow}
\usepackage{vcell}
\usepackage{booktabs,ctable,threeparttable}
\usepackage[cmex10]{amsmath}
\usepackage{dblfloatfix}
\usepackage{url}

\begin{document}
%
\title{Semantic Labeling of High Resolution Images Using EfficientUNets and Transformers}
%
%
%

\author{Hasan~AlMarzouqi,~\IEEEmembership{Senior Member,~IEEE,}
        and~Lyes~Saad~Saoud
\thanks{H. AlMarzouqi and L. Saad Saoud are with Electrical Engineering and Computer Science Department,Khalifa University, PO Box 127788, Abu Dhabi, UAE,e-mails:(hasan.amrzouqi, lyes.saoud)@ku.ac.ae}

\thanks{Manuscript received x xx, xxxx; revised x  xx, xxxx.}}

%
%

\markboth{Journal of \LaTeX\ Class Files,~Vol.~xx, No.~x, June~2022}%
{Shell \MakeLowercase{\textit{et al.}}: Bare Demo of IEEEtran.cls for Journals}
%



\maketitle

\begin{abstract}
Semantic segmentation necessitates approaches that learn high-level characteristics while dealing with enormous amounts of data. Convolutional neural networks (CNNs) can learn unique and adaptive features to achieve this aim. However, due to the large size and high spatial resolution of remote sensing images, these networks cannot analyze an entire scene efficiently. Recently, deep transformers have proven their capability to record global interactions between different objects in the image. In this paper, we propose a new segmentation model that combines convolutional neural networks with transformers, and show that this mixture of local and global feature extraction techniques provides significant advantages in remote sensing segmentation. In addition, the proposed model includes two fusion layers that are designed to represent multi-modal inputs and output of the network efficiently. The input fusion layer extracts feature maps summarizing the relationship between image content and elevation maps (DSM). The output fusion layer uses a novel multi-task segmentation strategy where class labels are identified using class-specific feature extraction layers and loss functions. Finally, a fast-marching method is used to convert unidentified class labels to their closest known neighbors. Our results demonstrate that the proposed methodology improves segmentation accuracy compared to state-of-the-art techniques.
\end{abstract}


%
\IEEEpeerreviewmaketitle

\section{Introduction}
%
%
%
%
\IEEEPARstart{I}{n} recent years, with the continual advancement of remote sensing technology, high-resolution remote sensing satellites have remarkably been utilized, and the resolution of remote sensing images has considerably improved \cite{WANG2022104969}. As a result, understanding detailed, high-resolution remote sensing images has become a significant challenge \cite{MA2019166}. Semantic image segmentation, also known as pixel-level categorization, is a critical computer vision challenge and is a vital technology for remote sensing image understanding \cite{MI2020140}. In semantic segmentation, the goal is to assign a class label to each image pixel \cite{YUAN2021114417}. 
There are two types of image semantic segmentation methods: conventional and deep-learning-based \cite{9081937}. Traditionally, machine learning approaches have used handmade features, whereas deep learning ones show higher performance by simultaneously learning feature representation and classifier parameters \cite{MI2020140}. 
Deep learning methods and particularly convolutional neural networks (CNN) were used successfully in solving multiple remote sensing problems. For example, a CNN-based model depending on a downsample-then-upsample architecture was used by Volpi et al. for semantic labelling of subdecimeter resolution images \cite{7725499}. An object-based classification technique was integrated with a deep learning model in \cite{7890382} to improve remote sensing image classification accuracy. Zhao et al. used CNNs to explore semantic segments, and a conditional random field was utilized to model the contextual information between them \cite{ZHAO201748}. A rotation equivariance CNN architecture was used for high-resolution land cover mapping in \cite{MARCOS201896}. Bergado et al. \cite{8388225} presented a single-stage approach that embeds the processing stages in a recurrent multiresolution convolutional network. A self-cascaded network was used to improve labeling coherence using a sequential global-to-local context aggregation method \cite{LIU201878}. Marmanis et al. \cite{MARMANIS2018158} presented an end-to-end trainable deep convolutional neural network for semantic segmentation with built-in knowledge of semantically important boundaries. Sun et al. \cite{SUN2019297} suggested ensemble techniques and a residual architecture for encoder-decoder models to mitigate the negative effects of structural stereotypes and address the issue of insufficient learning. Mi et al. \cite{MI2020140} used a differentiable decision forest for remote sensing semantic segmentation. Zhong et al. adapted the conventional FCN-8s/16s/32s models to extract roads and buildings from remote sensing RGB images \cite{7729406}.  Audebert et al. \cite{AUDEBERT201820} suggested and evaluated their FCN-based semantic segmentation approaches utilizing IRRG and DSM as input data. A residual dense U-Net was proposed for pixel-wise sea-land segmentation in complex and high-density remote sensing images \cite{8789636}. Symmetrical dense-shortcut U-Net was used to segment high-resolution remote sensing images \cite{doi:10.1080/01431161.2020.1742944}. The DeepLab semantic segmentation model and object-based image analysis were used to segment high resolution remote sensing images in \cite{doi:10.1080/17538947.2020.1831087}. Although the powerful representation capabilities of deep learning and CNN-based methods have aided in developing semantic segmentation of high-resolution images, achieved results remain far from optimal. 
Convolutional segmentation models rely on learnable convolutions that extract semantically significant features. Unfortunately, the local scope of convolutional filters restricts access to the relationships between distant image pixel intensities. Global features are particularly critical in remote sensing segmentation because the labeling of image patches is frequently dependent on the global context. To circumvent this limitation, DeepLab models \cite{doi:10.1080/17538947.2020.1831087} use dilated convolutions and spatial pyramid pooling. This enables the expansion of the receptive fields of convolutional networks and the extraction of multiscale information. Nevertheless, convolutional backbones remain biased toward local interactions, and fundamental changes in network architecture are required to solve this issue \cite{Strudel_2021_ICCV}. 
\begin{figure*}[]
\centering
    \includegraphics[width=1\linewidth,height=4.5cm]{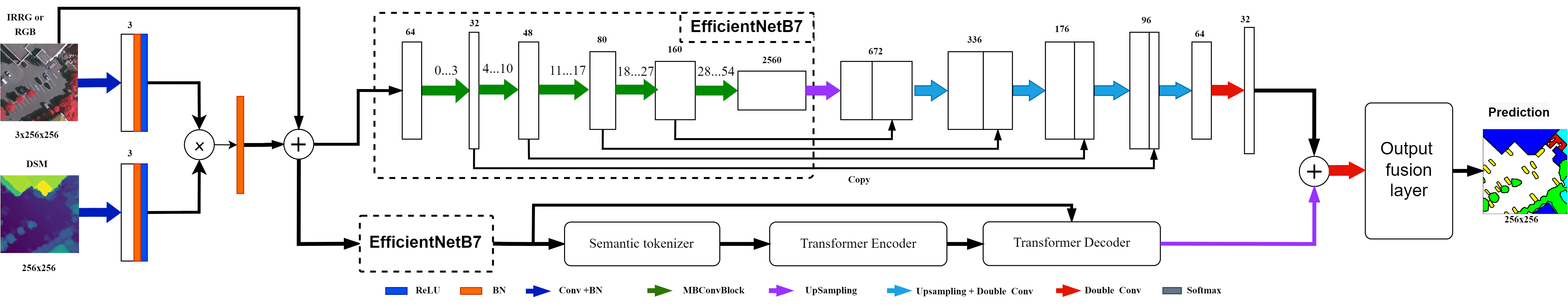}\
\caption{Overall model architecture. Double convolution layers apply the sequence conv.+BN+ReLU two times.}
\label{fig: 1}
\end{figure*}

Recently, transformer models have gained immense interest in solving computer vision tasks due to their effectiveness \cite{10.1145/3505244}. Transformers can record global interactions between elements in a scene. However, simulating global interactions has a quadratic cost, making such techniques prohibitively expensive computationally when applied to raw picture pixels. Despite the high performance shown by transformers in several computer vision problems, they have drawbacks that limit their capability. For example, transformers do not learn to attend locally in earlier layers, while incorporating local information at lower layers is vital for strong performance \cite{https://doi.org/10.48550/arxiv.2002.09402}.

Here, we propose an end-to-end fusion framework that combines U-Nets and transformers. Fig.~\ref{fig: 1} describes the architecture of the proposed ensemble model. The U-Net component models the dense connections between pixels, while the transformer-based component models the context using a token-based technique. 
local relationships between pixel intensities, while transformer features emphasize global interactions. 
The developed segmentation system consists of four major parts: an input fusion layer, a transformer-based network, an EfficientUNet network, and an output fusion layer.
In the input fusion layer, features representing the multi-modal input data: image content, and elevation maps (DSM) are extracted. The fused features are processed in parallel using the U-Net model and a transformer model. Finally, the resulting feature maps are passed through a novel multitask segmentation strategy that identifies class labels using class-specific feature extraction layers and loss functions. The use of these class-specific features and loss functions was shown to further improve the performance of the network. 

The contributions of this paper can be summarized as follows.
\begin{itemize}
  \item An efficient mixture model EfficientUNetTransformer is developed for the semantic segmentation of high-resolution remote sensing images. In this model, we combine transformers and the U-Net network to better represent global and local contexts, leading to more consistent labeling outcomes in complex urban constructions.
  \item	A novel multi-task segmentation approach that identifies class labels using class-specific feature extraction layers and loss functions
  \item	Extensive experiments on two publicly available datasets demonstrate the performance of the proposed model. The proposed ensemble model yields higher accuracy than the purely convolutional equivalent and outperforms several recently proposed attention-based semantic segmentation algorithms.
\end{itemize}
The remainder of this paper is organized as follows. Section II describes the proposed model in detail. Section III presents experimental results and discussions, while Section IV summarizes the conclusions.

\section{Proposed Model}
A flowchart of the proposed model is presented in Fig.~\ref{fig: 1}. The developed segmentation system consists of four major parts: 1) a transformer-based network that can extract global high-level semantic characteristics from the input image, 2) an EfficientUNet network that focuses on the extraction of local features from the input image, 3) The input fusion network that merges the IRRG or the RGB image with its digital surface model (DSM) image, and 4) the output fusion layer that splits the sum of local and global features into six separate binary sub classes by passing them through tokenizers and transformers. In what follows, we will provide a detailed description of the different components of the system. 

\subsection{Input fusion layer}
\begin{figure}  
\centering
\includegraphics[scale=0.35]{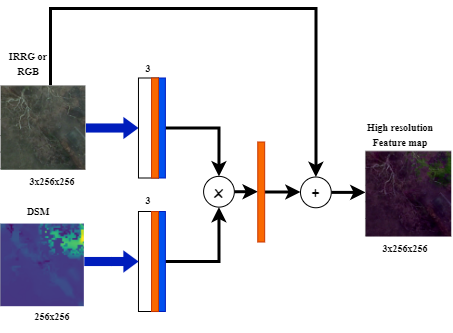}
\captionof{figure}{The input fusion layer. The white rectangle represents a convolution layer, orange: batch normalization, and blue: ReLU.}
\label{fig: 2}
\end{figure}

We propose a new input fusion layer that combines the IRRG or RGB image with the corresponding DSM image. This module passes both IRRG/RGB and DSM images through a sequence of convolution, batch normalization (BN), and ReLU blocks to extract high resolution features, then a dot multiplication is applied between the two obtained feature maps followed by BN layer. Finally, an addition operation is applied to the resulting feature map and the original IRRG/RGB image. These details are illustrated in Fig.~\ref{fig: 2}.

\subsection{Transformer path}
An EfficientNet B7 \cite{pmlr-v97-tan19a} deep neural network architecture is used to extract relevant features from image patches. First, we removed the last stage (i.e., the head stage) from the original EfficientNetB7, which originally contained nine stages. 
Next, input features are tokenized and sent to a transformer encoder. The tokenizer (Fig.~\ref{fig: 1}) takes the extracted features $X \in \Re ^{HxWxC}$, where H, W, and C are the input feature’s height, width, and channel dimension, and divides them into tokens $T =[T_1, T_2, \cdots, T_{N_f}]$, where $N_f$ is the size of the vocabulary set of tokens.
We use a point-wise convolution across the channel dimension to produce $N_f$ semantic groups for each pixel on EfficientNet features, with each group denoting one semantic idea (Fig.~\ref{fig: 3}). A softmax function is applied to the $HW$ dimension of each semantic group, to compute the feature maps $F$. Similarly, another point-wise convolution is used to produce $N_a$ semantic groups for each pixel in the EfficientNet features, with each group denoting one semantic idea (Fig. 3). We set the value of $N_a$ to equal the number of segmentation labels. A softmax function is applied to the $HW$ dimension of each semantic group to compute the attention layers $A$.

The semantic tokens $T_j$, $j=1, \cdots,{N_f}=32$ are computed using the following equation:
\begin{equation}
\label{eq:1}
\begin{aligned}
T_j = A^TF_j= [\sigma(\phi(A,W_1)]^{'}[\sigma(\phi(F_j,W_2)]
\end{aligned}
\end{equation}
where \(\phi(.)\) is a point-wise convolution with learnable kernels \( W_1 \in ~\Re ^{1x1xN_a}\)  and \(W_1 \in \Re ^{1x1xN_f}\) , and \(\sigma(.)\) is the softmax function. Each token $T_j$ is of size $H×W×6$.  EfficienetNet features maps are of size 65 × 65 × 32. The values of $N_a = 6$, and $N_f = 32$ are used in the experiments of this paper. 
The transformer encoder \cite{https://doi.org/10.48550/arxiv.2010.11929} is composed of encoders that translate a series of patch embeddings to pixel-level class labels. It has $L_D = 6$, layers of multi-head self-attention (MSA) and feedforward (FF) blocks (see Fig.~\ref{fig: 4}). At each layer $l$ , the input to self-attention is a triple (query Q, key K, value V) computed from the input $T^{(l-1)}$. Unlike the original transformer that uses the post-norm residual unit, we apply the layer normalization immediately before the MSA/MLP. The MSA unit can be described as: 
\begin{figure*}[]
\centering
    \includegraphics[width=0.9\linewidth]{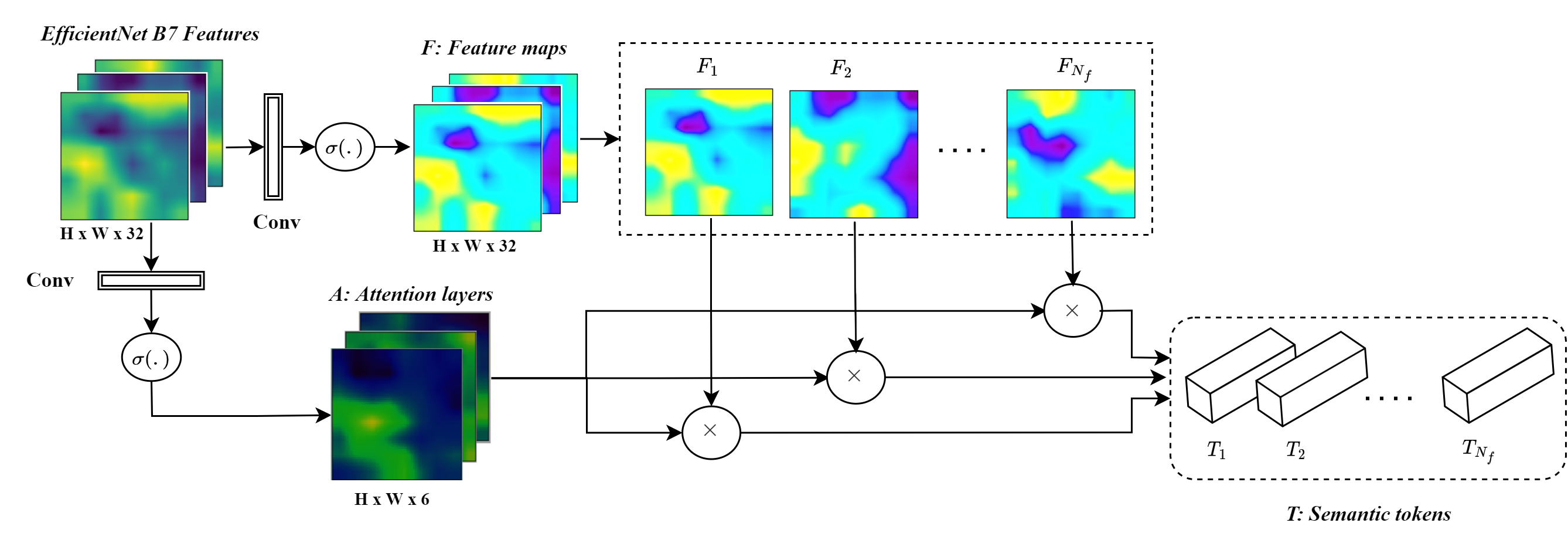}\
\caption{The semantic tokenizer. A point-wise convolution across the channel dimension extracts feature maps $F$ and attention layers A}
\label{fig: 3}
\end{figure*}
\begin{equation}
\label{eq:2}
\begin{aligned}
MSA(q,k,v) = Concat(head_1, \cdots, head_h)W^O 
\end{aligned}
\end{equation}
where
\begin{equation}
\label{eq:3}
\begin{aligned}
head_j = Att(qW_{j}^{q}, kW_{j}^{k},vW_{j}^{v} ) 
\end{aligned}
\end{equation}
where $qW_{j}^{q}$, $kW_{j}^{k}$,$vW_{j}^{v}$ are the linear projection matrices, and h is the number of attention heads.
The multi-head attention block receives three components—the query $Q$, the key $K$, and the value $V$—to compute the self-attention output
\begin{equation}
\label{eq:4}
\begin{aligned}
Attention(Q,K,V) = \sigma \left(\frac{QK'} {\sqrt{d} }\right)V
\end{aligned}
\end{equation}
where $d$ represents the channel dimension of the three components and $\sigma$ is the softmax function applied to the channel dimension. Finally, the output of the transformer is upsampled to match the dimension of features extracted from the EfficientUNet network. 
The transformer decoder comprises $L_D=6$ layers of multi-head cross attention (MCA) and FF blocks. The encoder's patch-level encodings are mapped to patch-level class scores by the decoder. The decoder and encoder configurations are equivalent. The MCA receives the query from the extracted features $X$, the key, and the value from the tokens $T$ generated by the transformer encoder (Fig.~\ref{fig: 4}).  
\subsection{EfficientUNet}
The second part of the proposed model is U-Net segmentation model that uses an architecture from the EfficientNet family of networks as a backbone. EfficientNet image classification models apply a compound-scaling approach that consistently adjusts the network depth, width, and resolution for increased performance using a given set of scaling parameters \cite{pmlr-v97-tan19a}. Scaling the network incrementally increases model performance by balancing the architecture's breadth, depth, and image resolution compound coefficients. EfficientNet is built using mobile inverted bottleneck convolution (MBConv), as shown in Fig.~\ref{fig: 5}b. The proposed model uses the swish activation function \cite{Sandler_2018_CVPR} instead of the widely used rectifier linear units (ReLUs). When going from EfficientNetB0 to EfficientNetB7, the depth, width, resolution, and model size increase, while the accuracy improves \cite{pmlr-v97-tan19a}. The architecture used in the proposed model is the EfficientNetB7 which has 55 basic building MBConv blocks as shown in Fig.~\ref{fig: 5}a. The components used in these blocks are shown in Fig.~\ref{fig: 5}b.
The proposed U-Net encoder is based on the EfficientNetB7 model. The decoder is constructed using a reversed version of the encoder model with upsampling units. The encoder outputs of layers 3, 10, 17, 27 are concatenated with their corresponding decoder outputs, as shown in Fig.~\ref{fig: 1}. Transposed convolution layers were employed to build the decoder, which doubled the size of a feature map while decreasing the number of channels by half. An upsampling layer followed by Double convolution layers were applied after each concatenation operation. Double convolution layers apply the sequence of convolution, batch normalization, and ReLU operations two times. 
\subsection{Output Fusion layer} 
This module is composed of a CNN that receives as input the sum of the final feature mappings from the two deep networks: the transformer and EfficientUNet. Input feature maps were summed and fed into a shallow CNN, which is a Double convolution layer with 32 input channels and 6 output channels.  The CNN’s output was sent to six separate tokenizers plus transformer encoders, representing the six binary classes, and then passed through logarithmic softmax functions (Fig.~\ref{fig: 6}).  
\begin{figure}  
\centering
\includegraphics[scale=0.18]{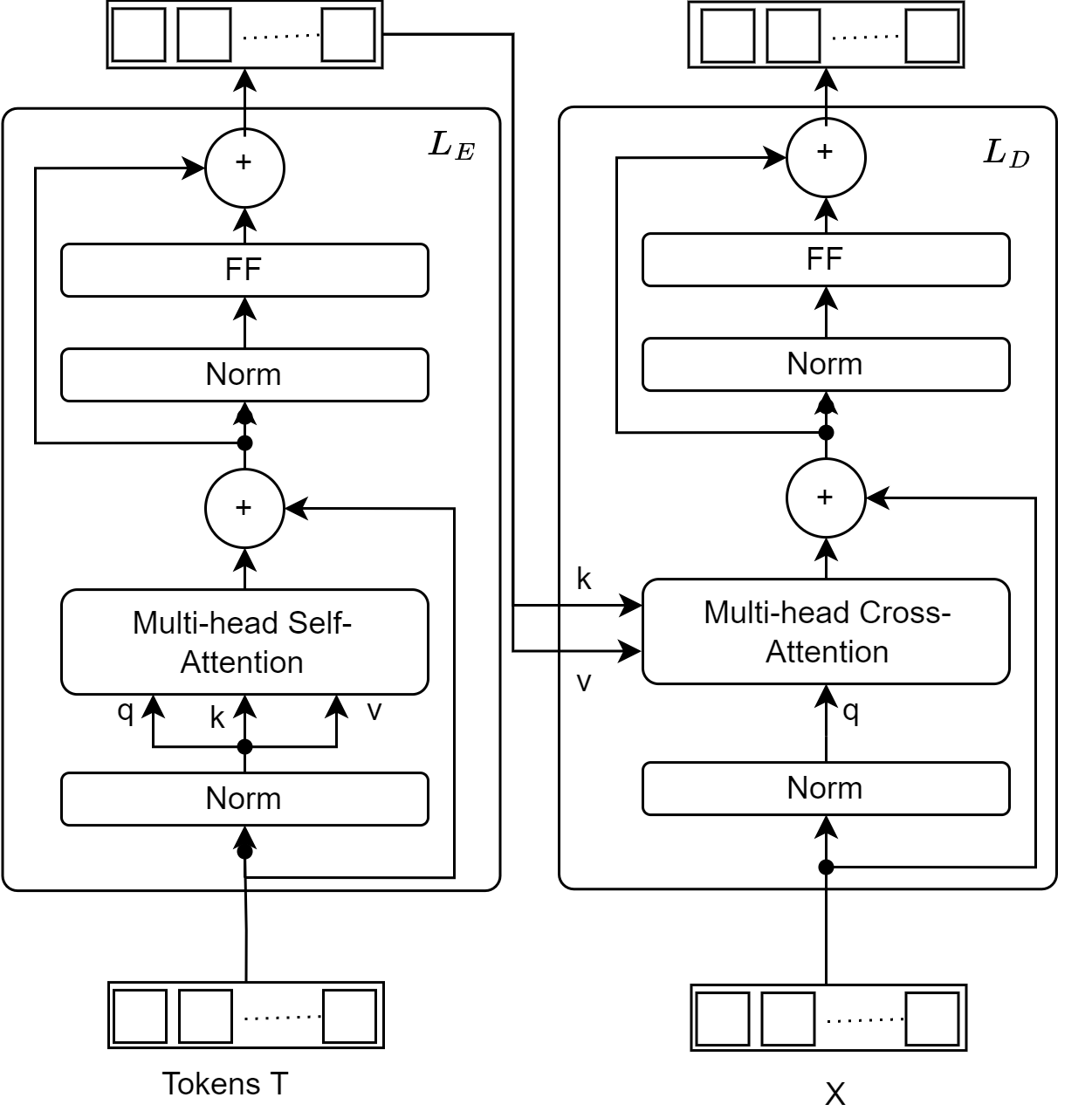}
\captionof{figure}{The transformer encoder and decoder}
\label{fig: 4}
\end{figure}
\begin{figure}[hptb]
\begin{subfigure}{.18\textwidth}
   \includegraphics[width=\textwidth]{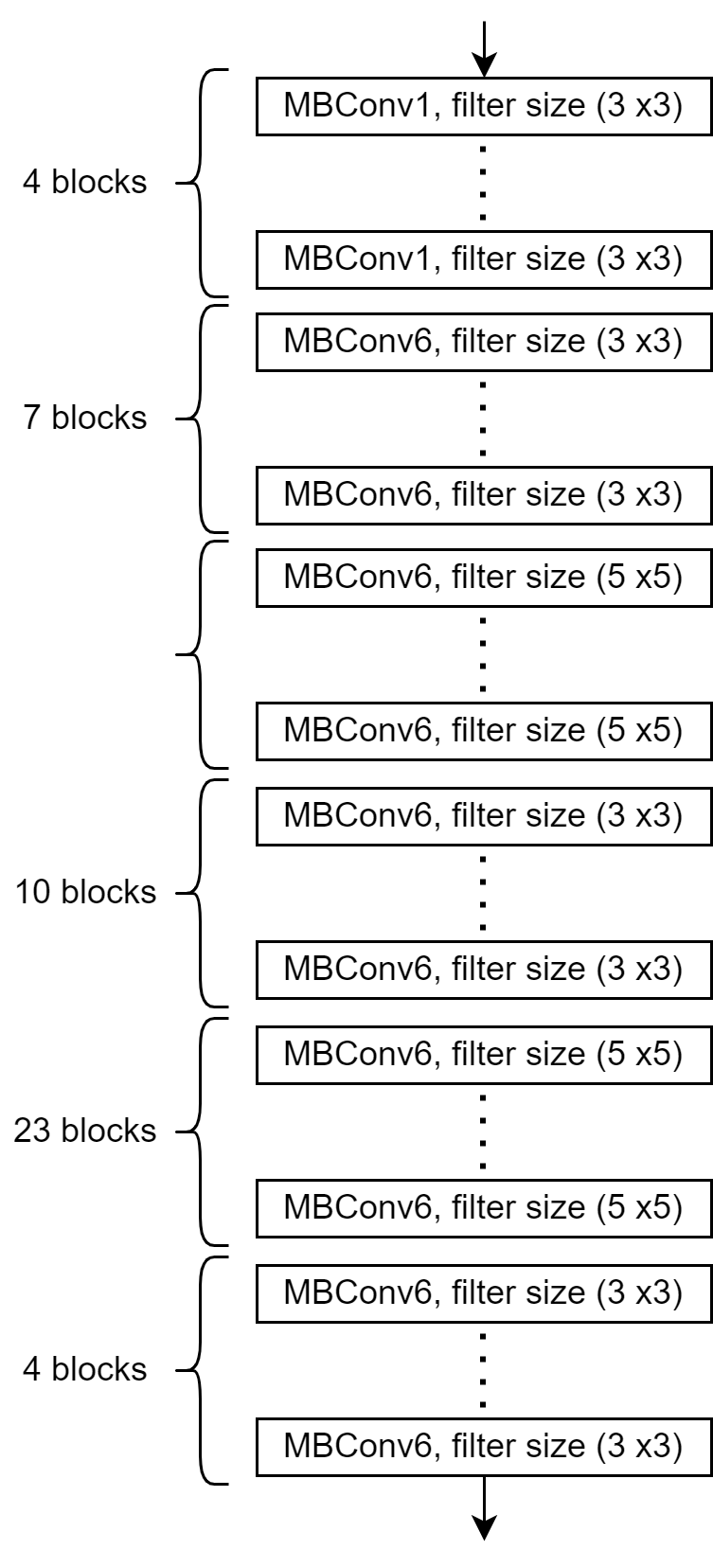}
   \caption{(a)}
\label{fig:g1}
\end{subfigure}%
\hfill
\begin{subfigure}{.27\textwidth}
   \includegraphics[width=\textwidth]{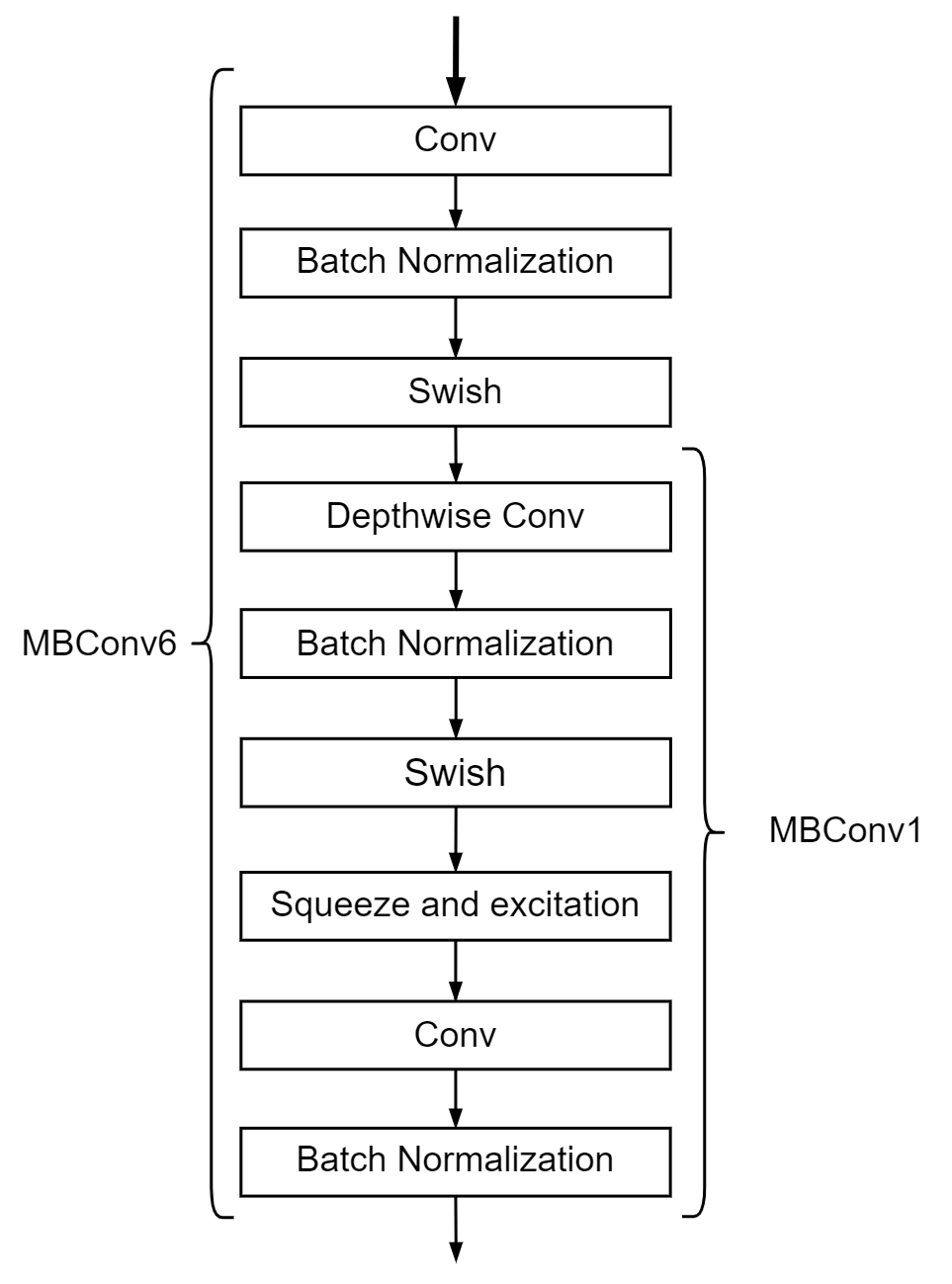}
   \caption{(b)}
\label{fig:g2}
\end{subfigure}
\caption{(a) EfficietnetB7 architecture (b) The mobile inverted bottleneck convolution layer (MBConv)}
\label{fig: 5}
\end{figure}
\begin{figure*}  
\centering
\includegraphics[width=0.7\textwidth]{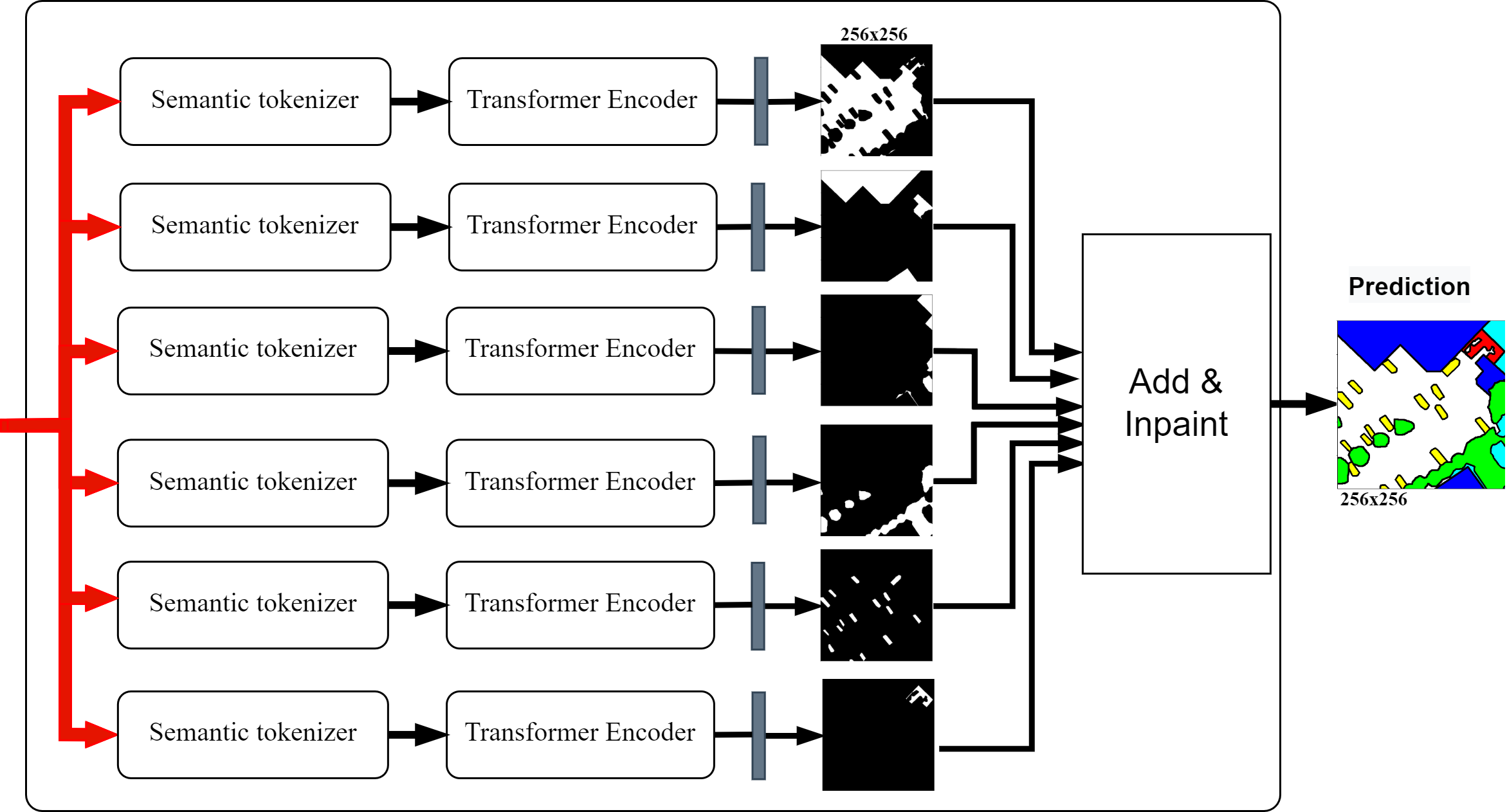}
\captionof{figure}{Output fusion layer}
\label{fig: 6}
\end{figure*}
Next, the logistic loss was used as the loss function and was computed using a Logarithmic Softmax layer and averaged across the entire patch \cite{AUDEBERT201820}:
\begin{equation}
\label{eq:5}
\begin{aligned}
loss = \frac{1}{N} {\sum_{i=1}^{N} {y^i log\left( \frac{exp\left(\hat{y}^i\right)}{\sum_{l=1}^{k} {exp\left(\hat{y}_{l}^{i}\right)}} \right)} } 
\end{aligned}
\end{equation}

where $N$ is the number of pixels in the input image, $k$ equals 2 classes, and for each pixel $i$ , $y^i$  and  $\hat{y}^i$ are the true and predicted labels, respectively. The different parts of the networks were trained together to find the best model that can predict the right labels without using special pre- or post-processing operations. 
The semantic segmentation maps are obtained by combining the six binary classes using the Add\&Inpaint layer (Fig.~\ref{fig: 6}). This layer includes all classes computed in the output fusion layer in the output segmentation map and then replaces misclassified pixels by their nearest classified neighbor using a fast-marching method (FMM) \cite{doi:10.1080/10867651.2004.10487596}. 
\section{Experimental Results and Discussions} 
\textbf{Dataset Description:} The proposed EfficientUNet model was evaluated on two challenging public datasets for semantic labeling.

\textbf{a)	ISPRS Vaihingen Challenge Dataset:} This dataset served as the baseline for the ISPRS 2D Semantic Labeling Challenge in Vaihingen \cite{labeling_2016}. It consists of three bands of infrared, red, and green (IRRG) image data and digital surface model (DSM), and normalized digital surface model (NDSM) data \cite{Gerke2014UseOT}. In all, there are 33 images with a ground sampling distance of 9 cm in the image data. We divided the 16 pictures with available ground truth into 12 images for a training set and 4 images ("5", "21", "15", and "30") for the validation. We utilized all 16 images as the training set for the test.

\textbf{b) ISPRS Potsdam Challenge Dataset:} The second dataset used in this study belongs to the Potsdam ISPRS 2D Semantic Labeling Challenge \cite{labeling_2016}. It is made up of four-band infrared, red, green, and blue (IRRGB) image data and matching DSM and NDSM data. Of the 38 images of 5{-}cm resolution, 24 images had the ground truth available, while the remaining 14 were kept by the challenge organizer for testing. From the 24 images given by the challenge organizer, we selected 17 images for training and 7 images ({"}3{\_}11{"}, {"}3{\_}12{"}, {"}4{\_}11{"},{"}5{\_}10{"}, {"}6{\_}9{"}, {"}6{\_}12{"}, {"}7{\_}11{"}) for validation. We utilized all 24 images as a training set for the test. 
In both datasets, patches of size 256 × 256 pixels were extracted from the images in the dataset using a sliding window with a stride value of 32.  

Proposed system was implemented in PyTorch. All of our models were trained using stochastic gradient descent (SGD) with a base learning rate of 0.01, momentum of 0.9, weight decay of 0.0005, and batch size of 10. The encoder-decoder weights were randomly initialized. We divided the learning rate by 10 after 25 and 45 epochs (out of a total of 100 epochs used for training). 
We present the overall pixel-wise accuracy (OA), the average F1 score, and the Cohen's kappa coefficient $\kappa$ across all classes to quantitatively evaluate performance. The F1 score and kappa coefficient $\kappa$ for a class $i$  are defined as follows:
\begin{equation}
\label{eq:6}
\begin{aligned}
F1_i = 2 \frac{tp_i}{C_i+P_i} 
\end{aligned}
\end{equation}
\begin{equation}
\label{eq:7}
\begin{aligned}
\kappa = \frac{p_o-p_e}{1-p_e} 
\end{aligned}
\end{equation}
where $tp_i$ is the number of true positives for class $i$, $C_i$ is the number of pixels in class $i$, $P_i$is the number of pixels assigned to class $i$ by the model, $p_o$ is the relative observed agreement among raters, and $p_e$ is the hypothetical probability of chance agreement. In compliance with the competition organizers' assessment guidelines, these metrics were derived after eroding the boundaries with a three-pixel radius circle and deleting those pixels \cite{AUDEBERT201820}.

\textbf{Performance comparison}
We compared the performance of the proposed model against several recently proposed models using validation and testing datasets. 

\textbf{On Validation data:} For both datasets, the proposed model was compared against the following five deep learning models: Fully connected networks (FCN-8s) \cite{Long_2015_CVPR}, U-Net \cite{8309343}, SegNet \cite{7803544}, PSPNet \cite{8100143}, and deep transformers \cite{9491802}.  

\begin{figure*}[hptb]
\begin{subfigure}{.11\textwidth}
   \includegraphics[width=\textwidth]{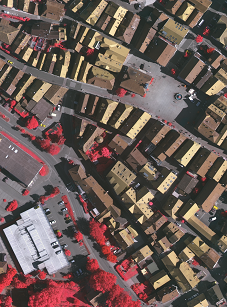}
   \caption{}
\end{subfigure}%
\hfill
\begin{subfigure}{.11\textwidth}
   \includegraphics[width=\textwidth]{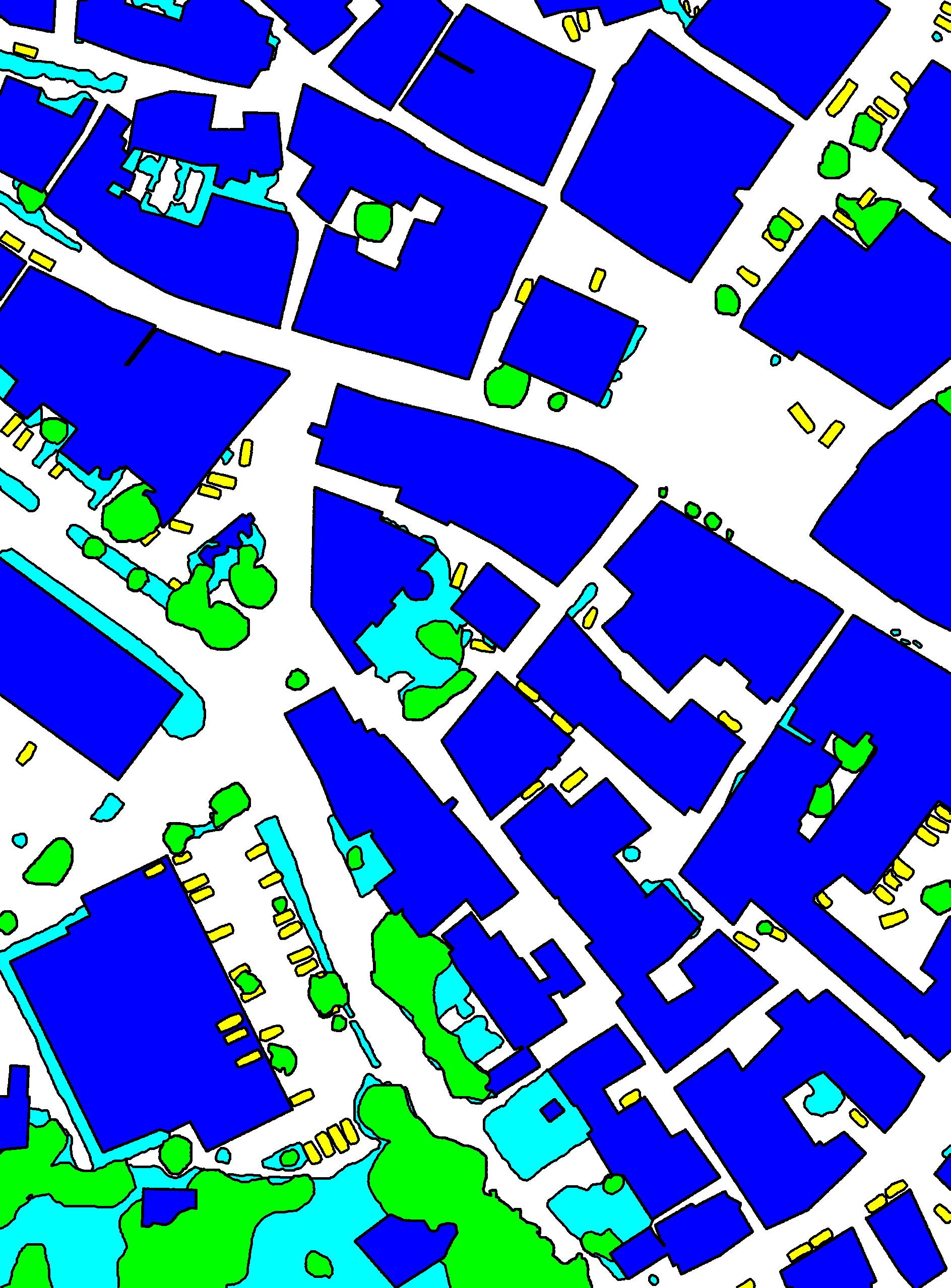}
   \caption{}
\end{subfigure}
\hfill
\begin{subfigure}{.11\textwidth}
   \includegraphics[width=\textwidth]{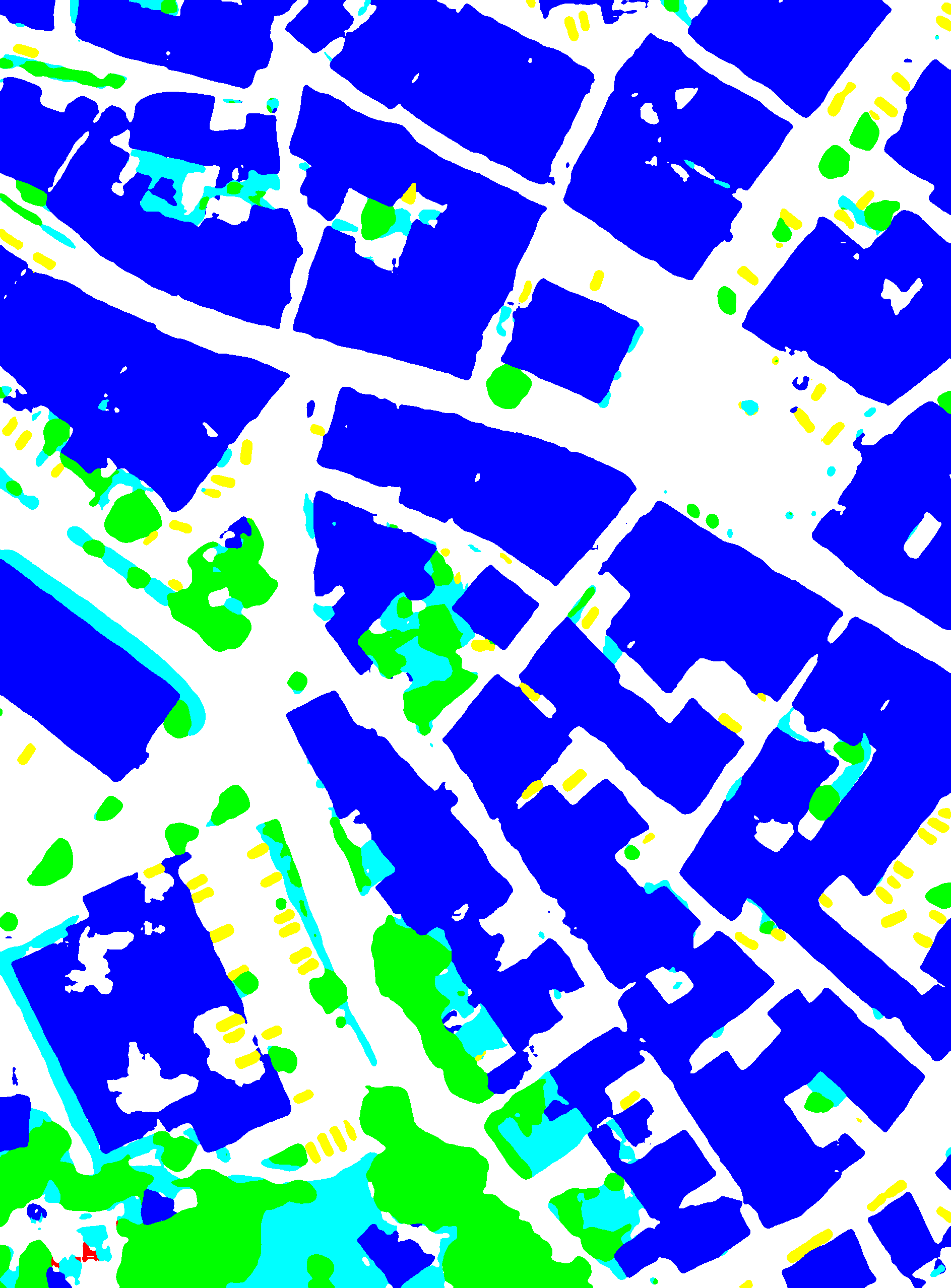}
   \caption{}
\end{subfigure}
\hfill
\begin{subfigure}{.11\textwidth}
   \includegraphics[width=\textwidth]{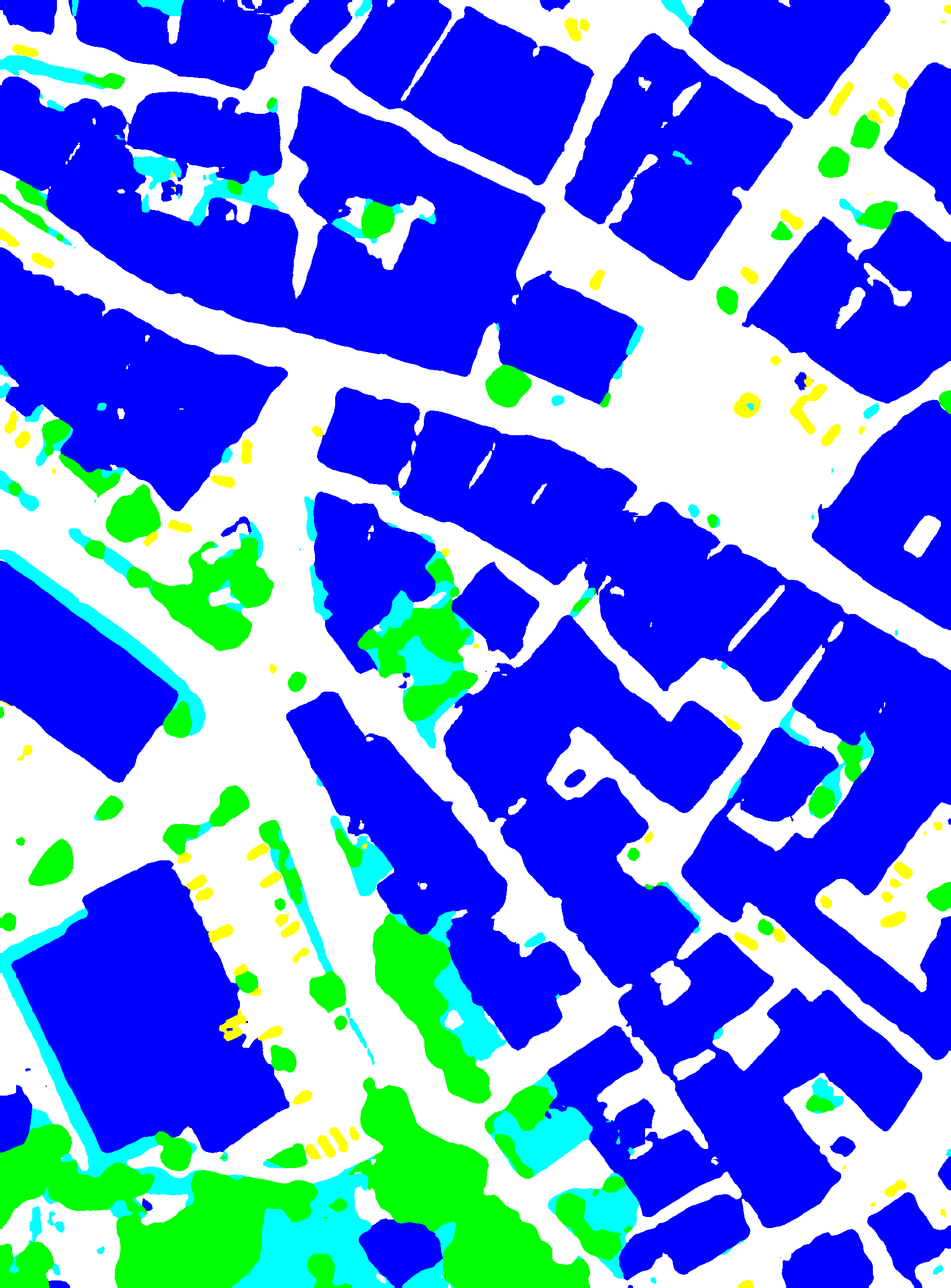}
   \caption{}
\end{subfigure}
\hfill
\begin{subfigure}{.11\textwidth}
   \includegraphics[width=\textwidth]{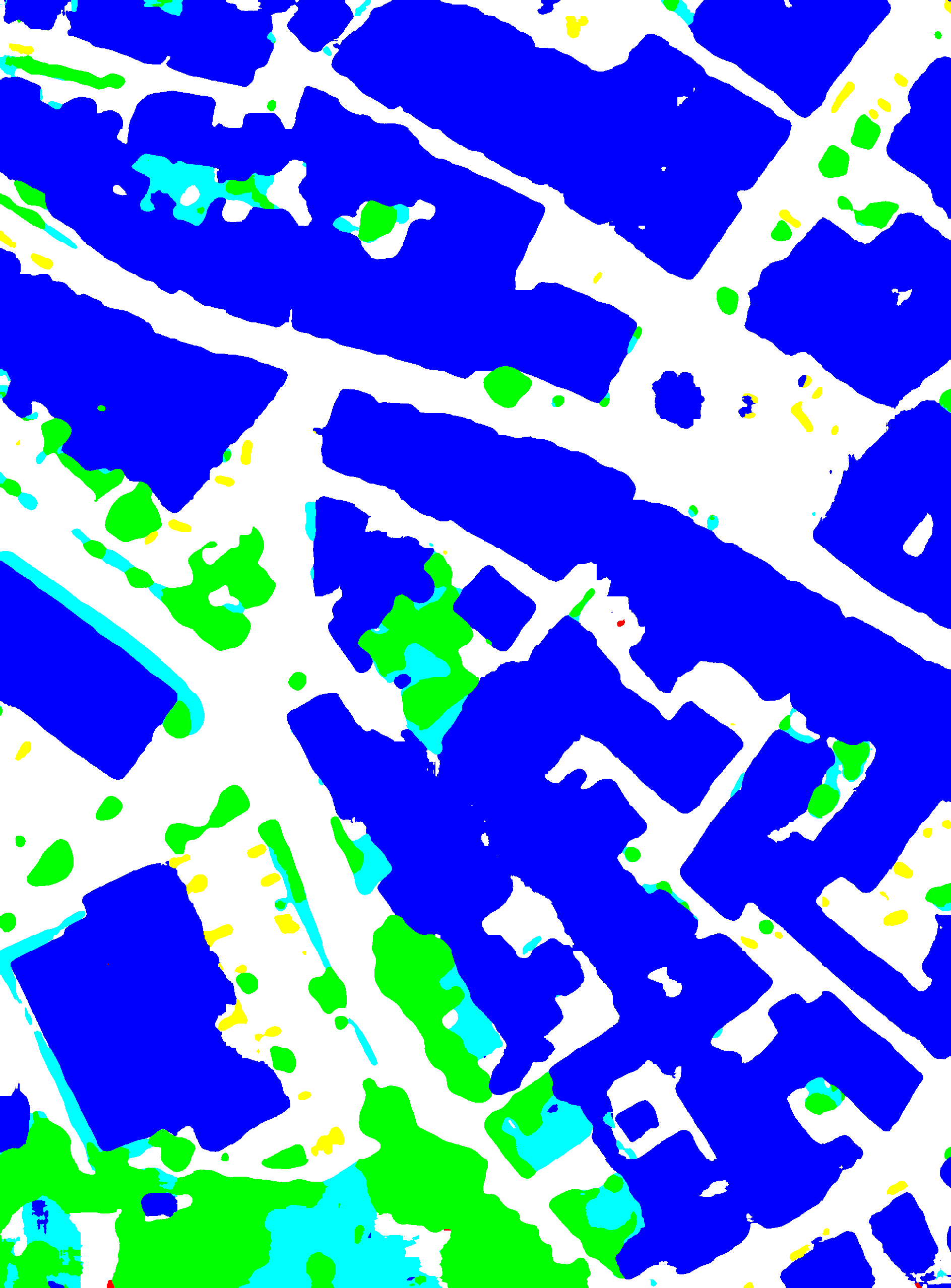}
   \caption{}
\end{subfigure}
\hfill
\begin{subfigure}{.11\textwidth}
   \includegraphics[width=\textwidth]{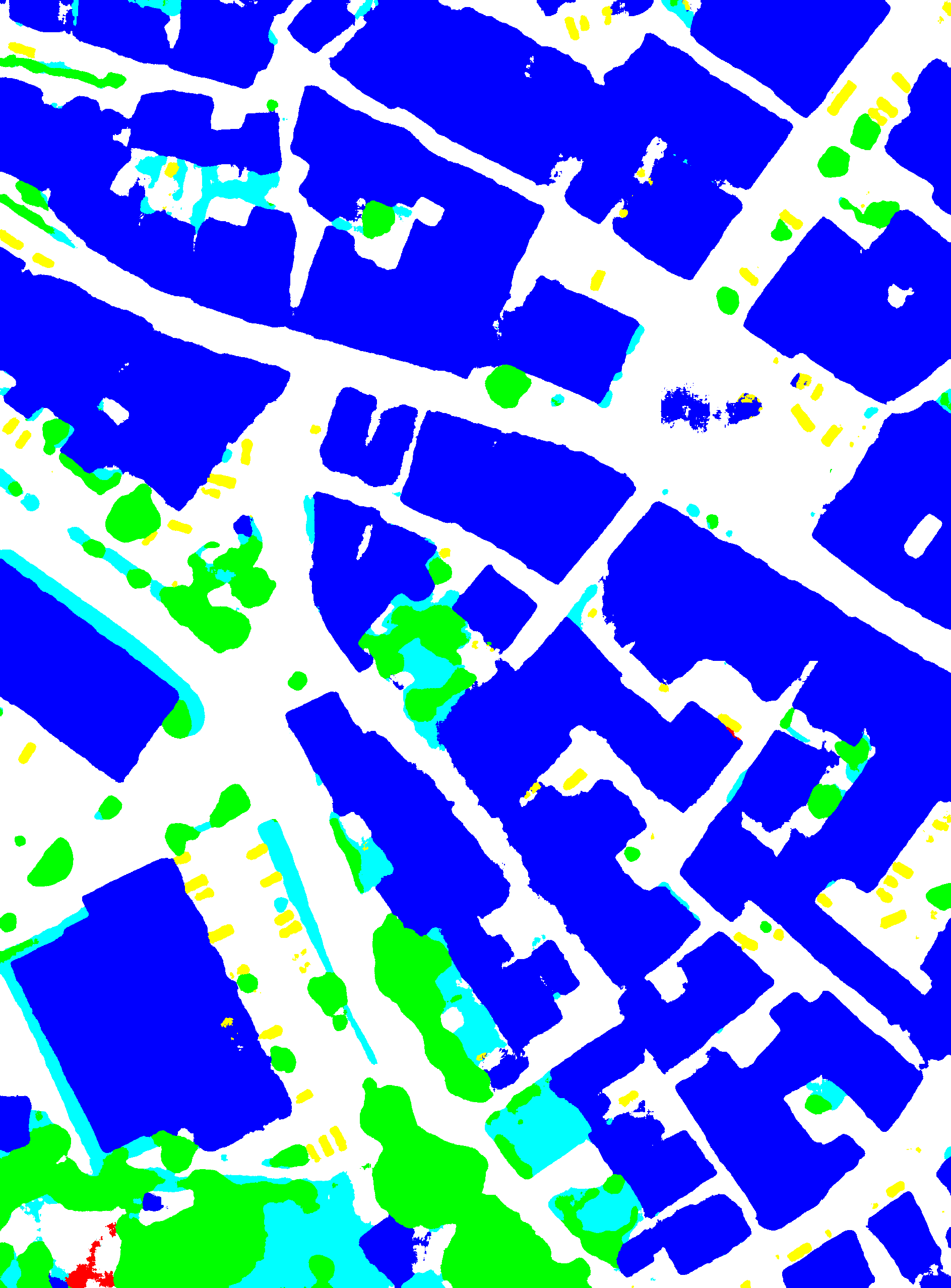}
   \caption{}
\end{subfigure}
\hfill
\begin{subfigure}{.11\textwidth}
   \includegraphics[width=\textwidth]{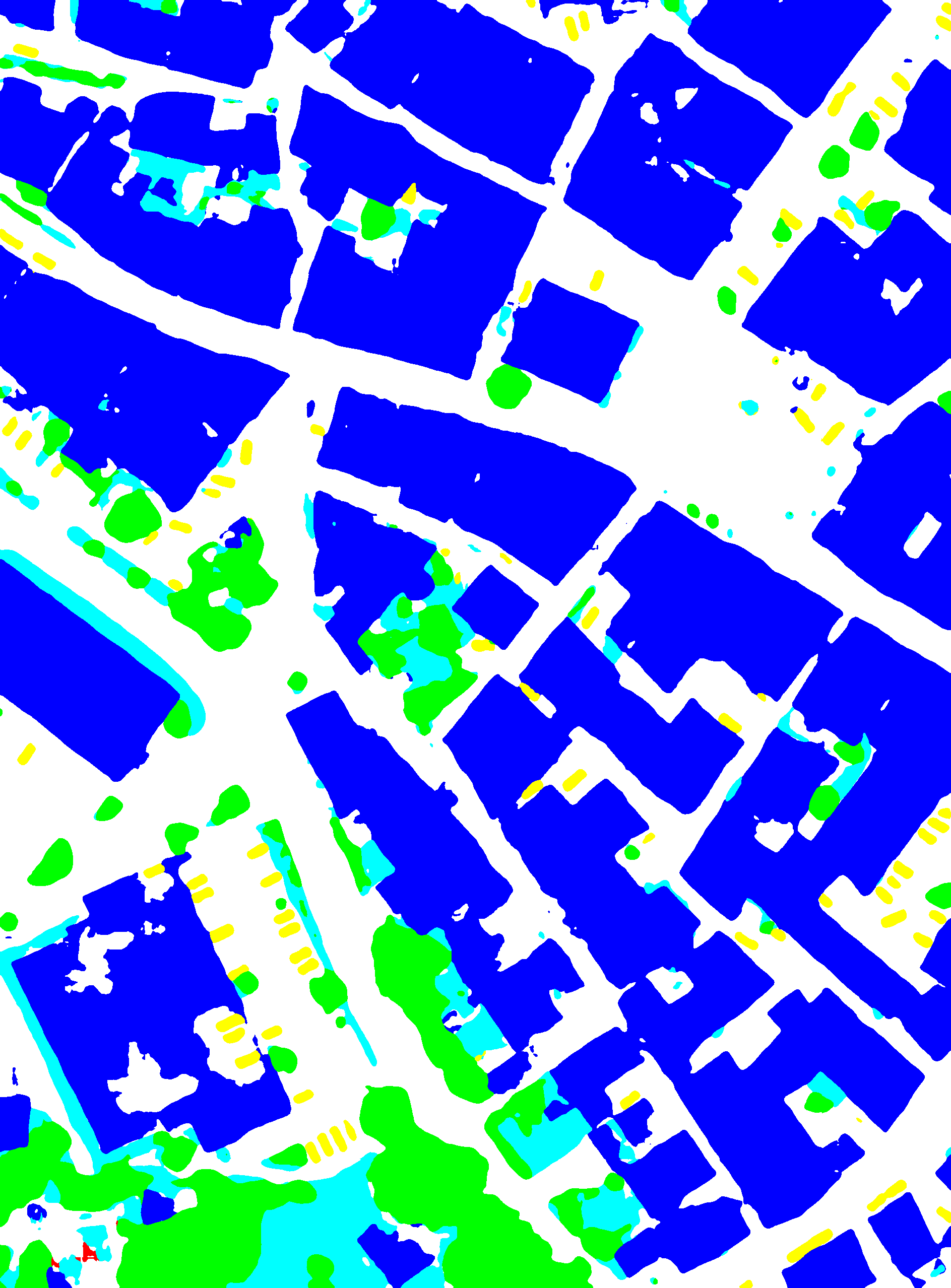}
   \caption{}
\end{subfigure}
\hfill
\begin{subfigure}{.11\textwidth}
   \includegraphics[width=\textwidth]{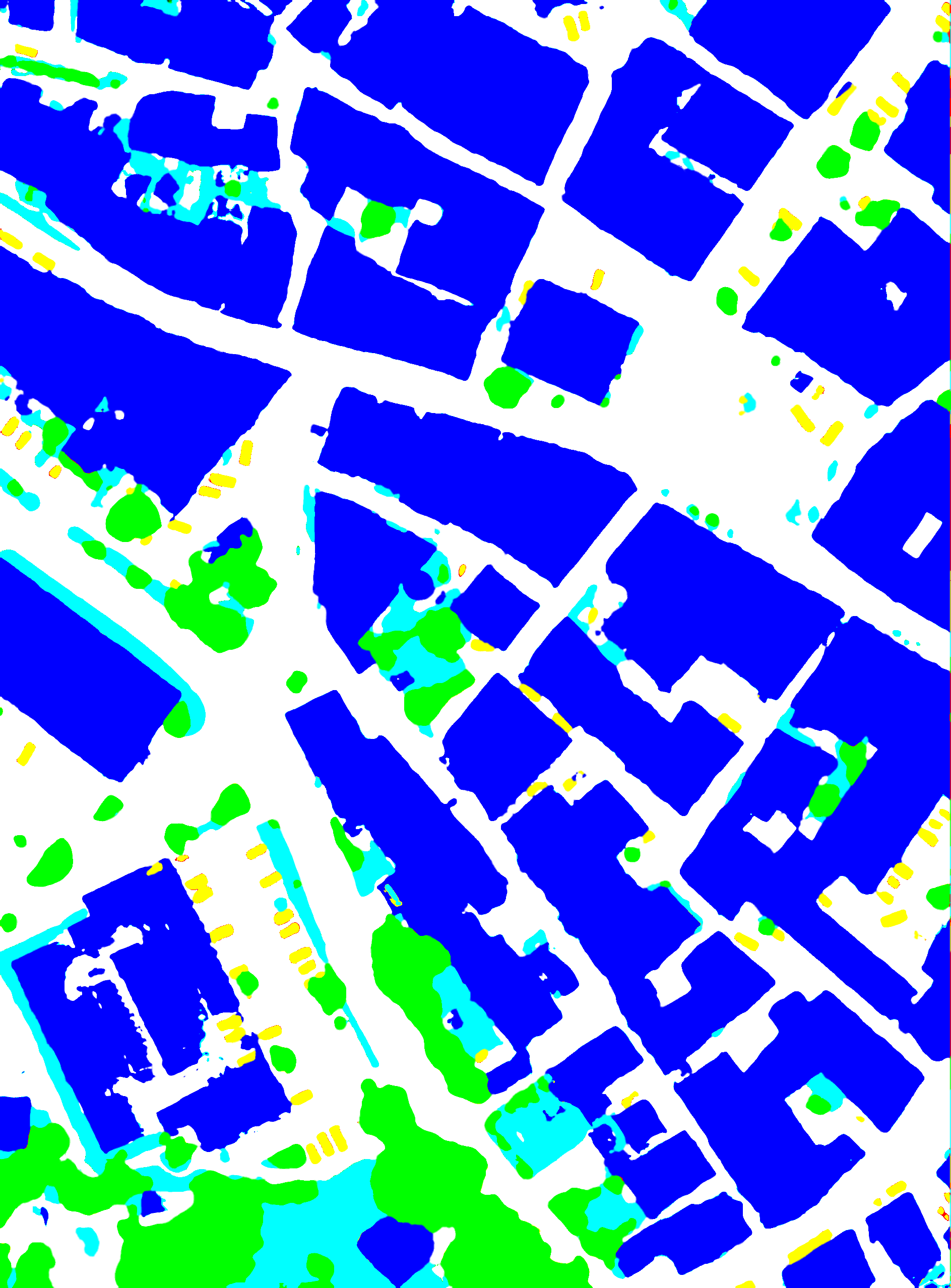}
   \caption{}
\end{subfigure}
\vfill
\begin{subfigure}{.11\textwidth}
   \includegraphics[width=\textwidth]{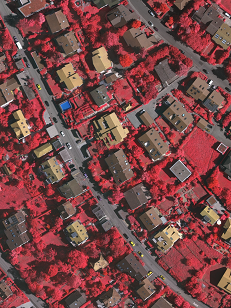}
   \caption{IRRG}
\end{subfigure}
\hfill
\begin{subfigure}{.11\textwidth}
   \includegraphics[width=\textwidth]{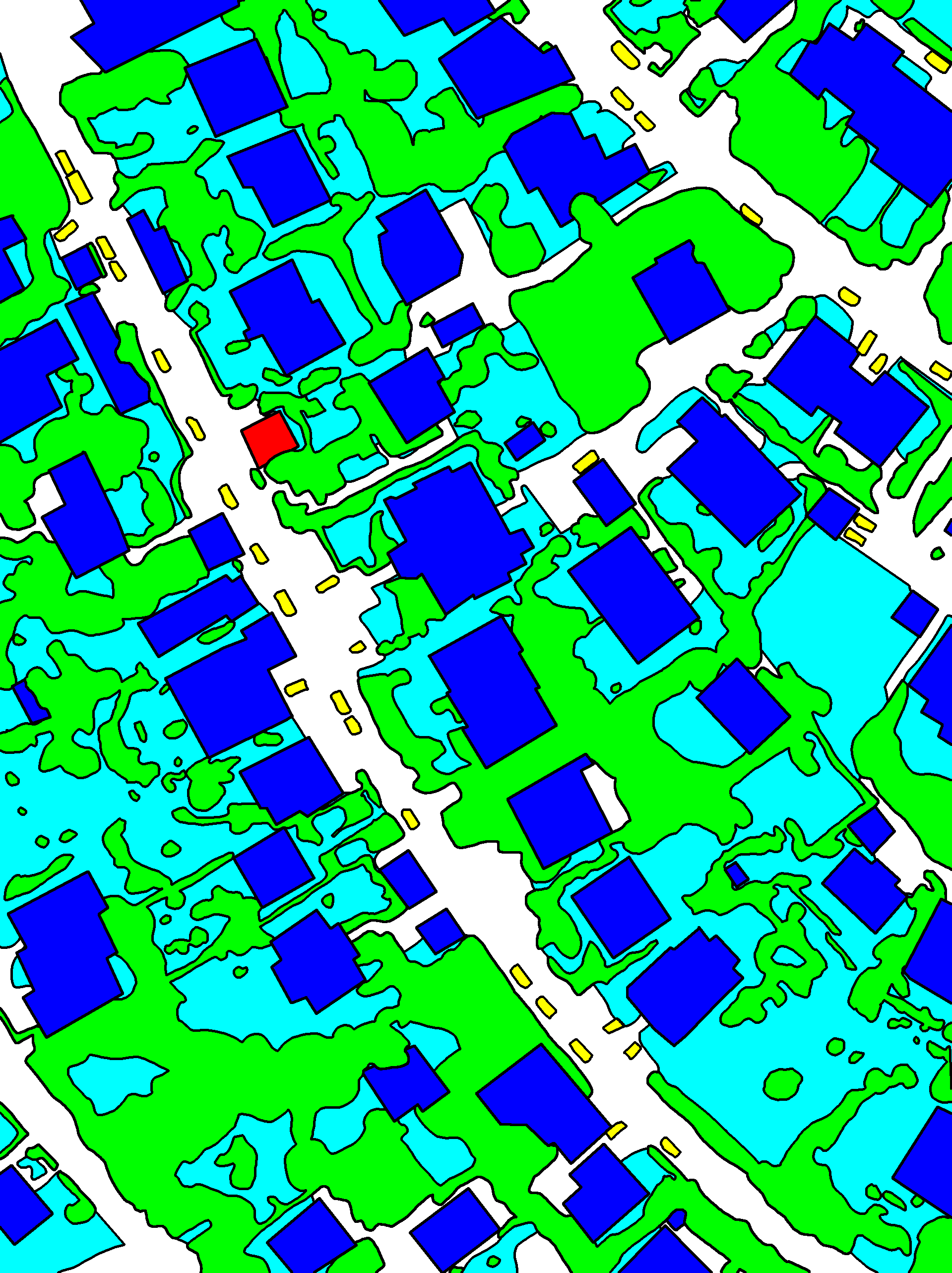}
   \caption{Ground truth}
\end{subfigure}
\hfill
\begin{subfigure}{.11\textwidth}
   \includegraphics[width=\textwidth]{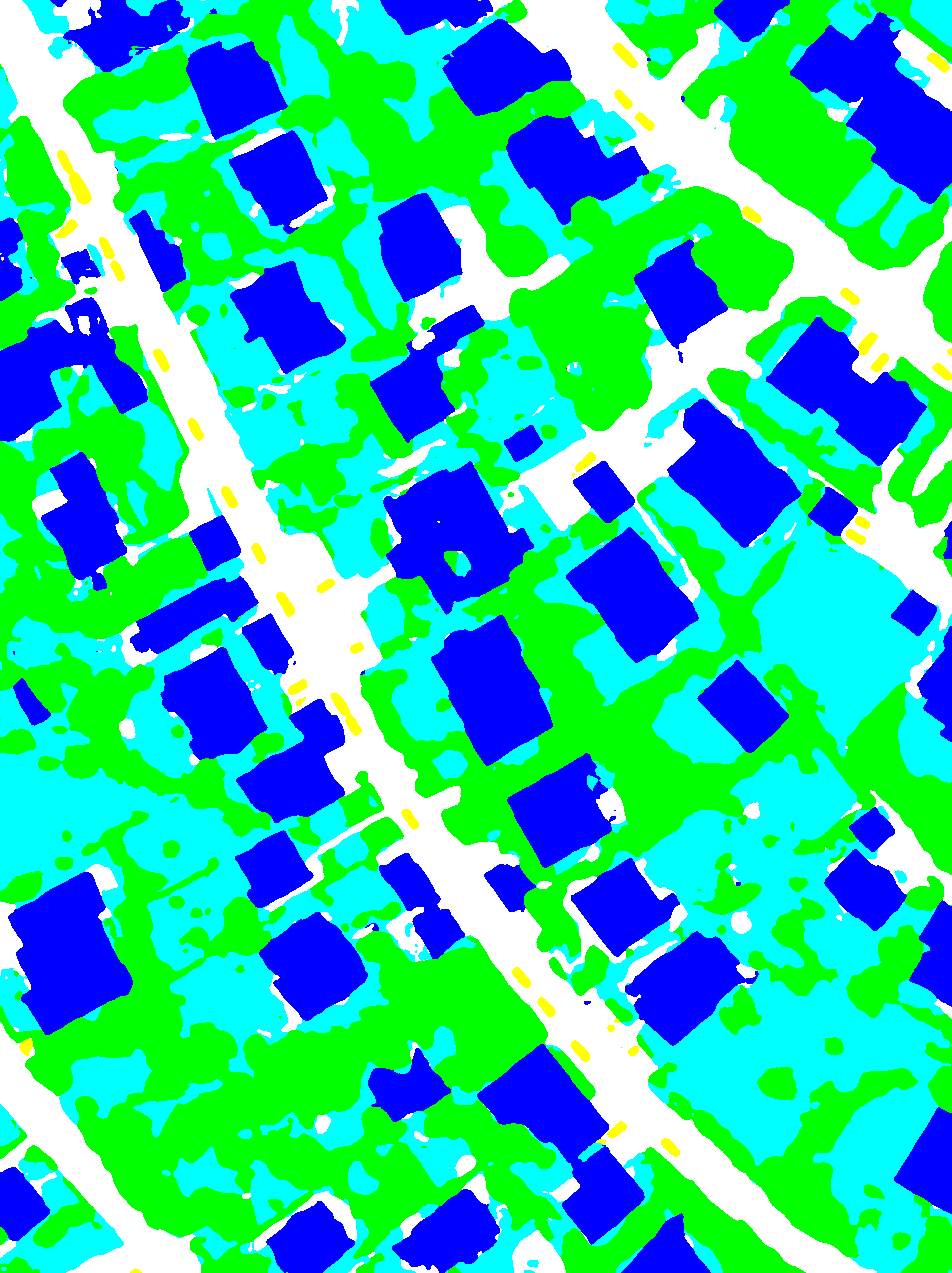}
   \caption{FCN8s}
\end{subfigure}
\hfill
\begin{subfigure}{.11\textwidth}
   \includegraphics[width=\textwidth]{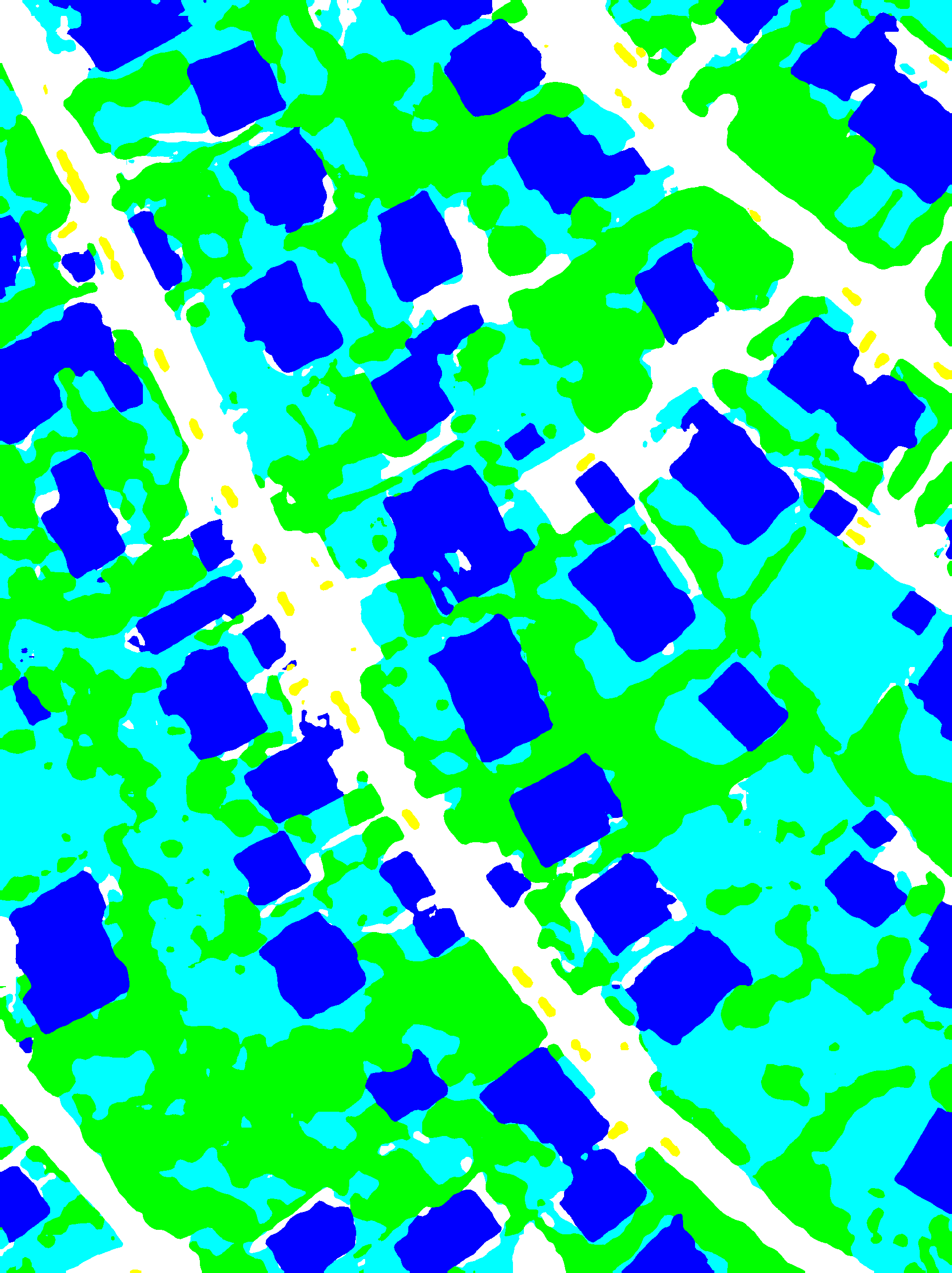}
   \caption{PSPNet}
\end{subfigure}
\hfill
\begin{subfigure}{.11\textwidth}
   \includegraphics[width=\textwidth]{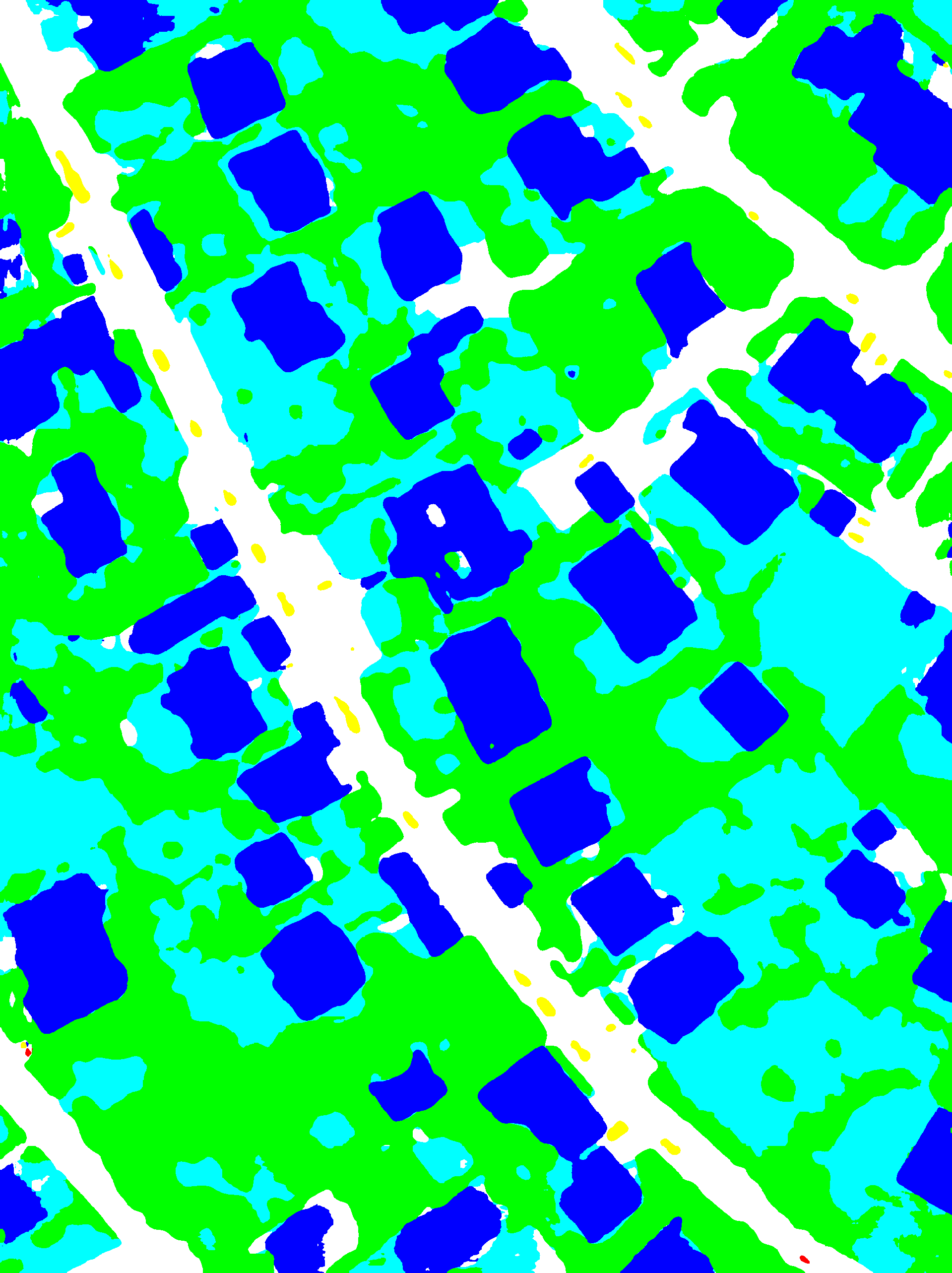}
   \caption{Unet}
\end{subfigure}
\hfill
\begin{subfigure}{.11\textwidth}
   \includegraphics[width=\textwidth]{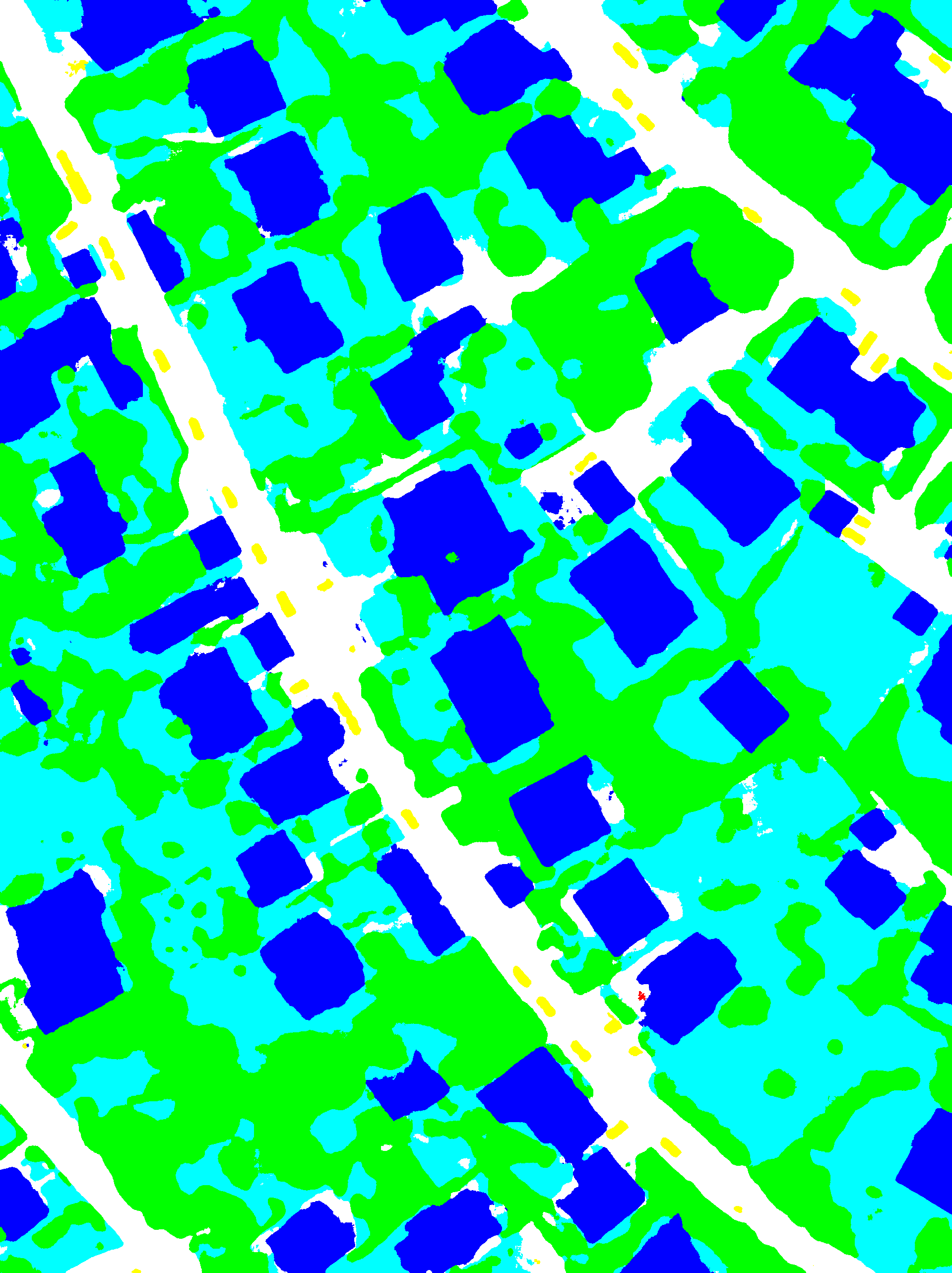}
   \caption{SegNet}
\end{subfigure}
\hfill
\begin{subfigure}{.11\textwidth}
   \includegraphics[width=\textwidth]{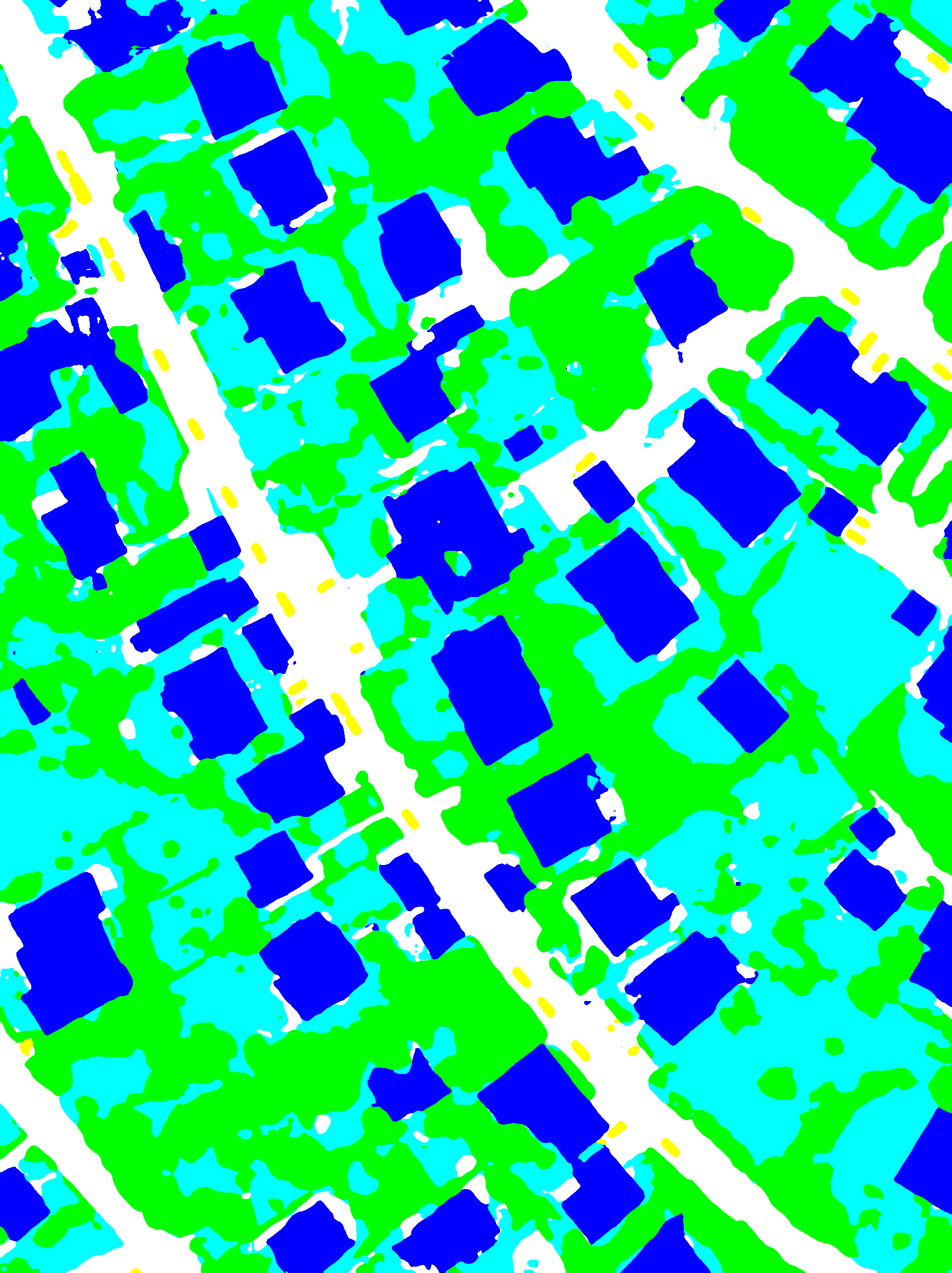}
   \caption{Transformer}
\end{subfigure}
\hfill
\begin{subfigure}{.11\textwidth}
   \includegraphics[width=\textwidth]{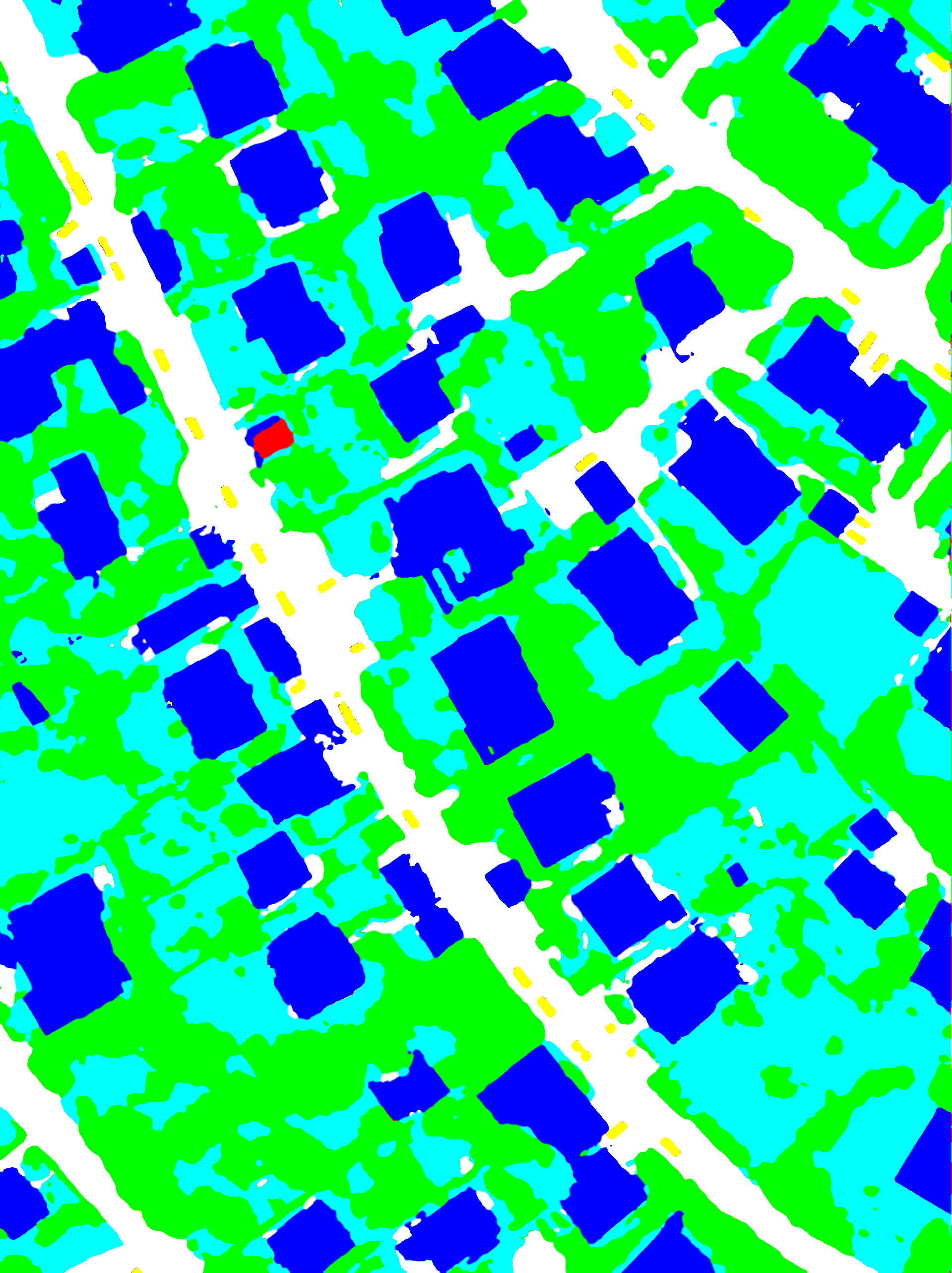}
   \caption{Proposed}
\end{subfigure}

\caption{Performance comparison with other deep learning models using the Vaihingen validation dataset. The labels represent the impervious surface in white, buildings in blue, low vegetation in cyan, trees in green, cars in yellow, and clutter/background in red.}
\label{fig: 7}
\end{figure*}

\begin{figure*}[hptb]
\begin{subfigure}{.11\textwidth}
   \includegraphics[width=\textwidth]{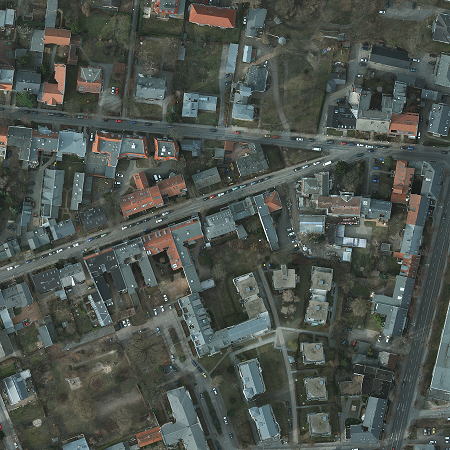}
   \caption{}
\end{subfigure}
\hfill
\begin{subfigure}{.11\textwidth}
   \includegraphics[width=\textwidth]{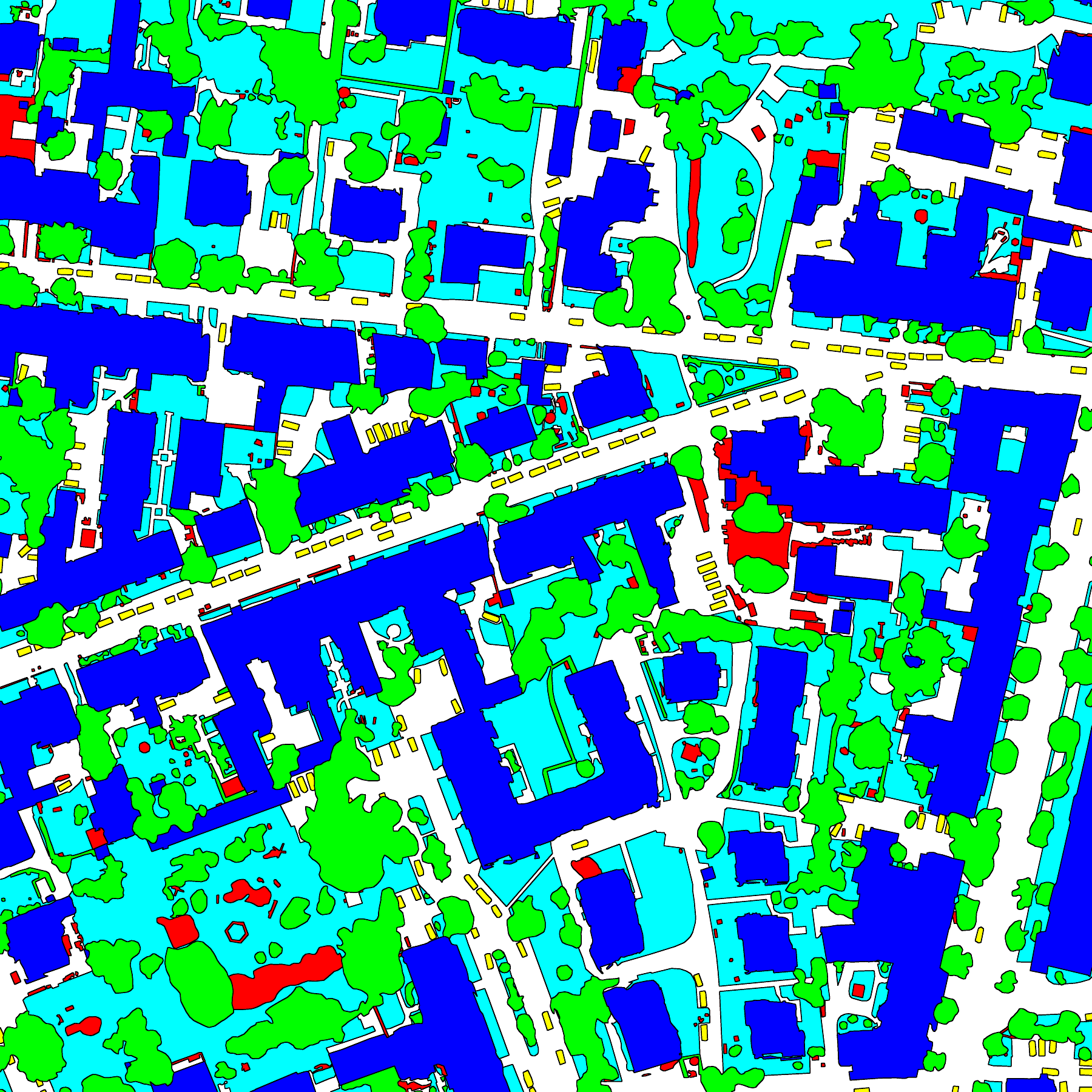}
   \caption{}
\end{subfigure}
\hfill
\begin{subfigure}{.11\textwidth}
   \includegraphics[width=\textwidth]{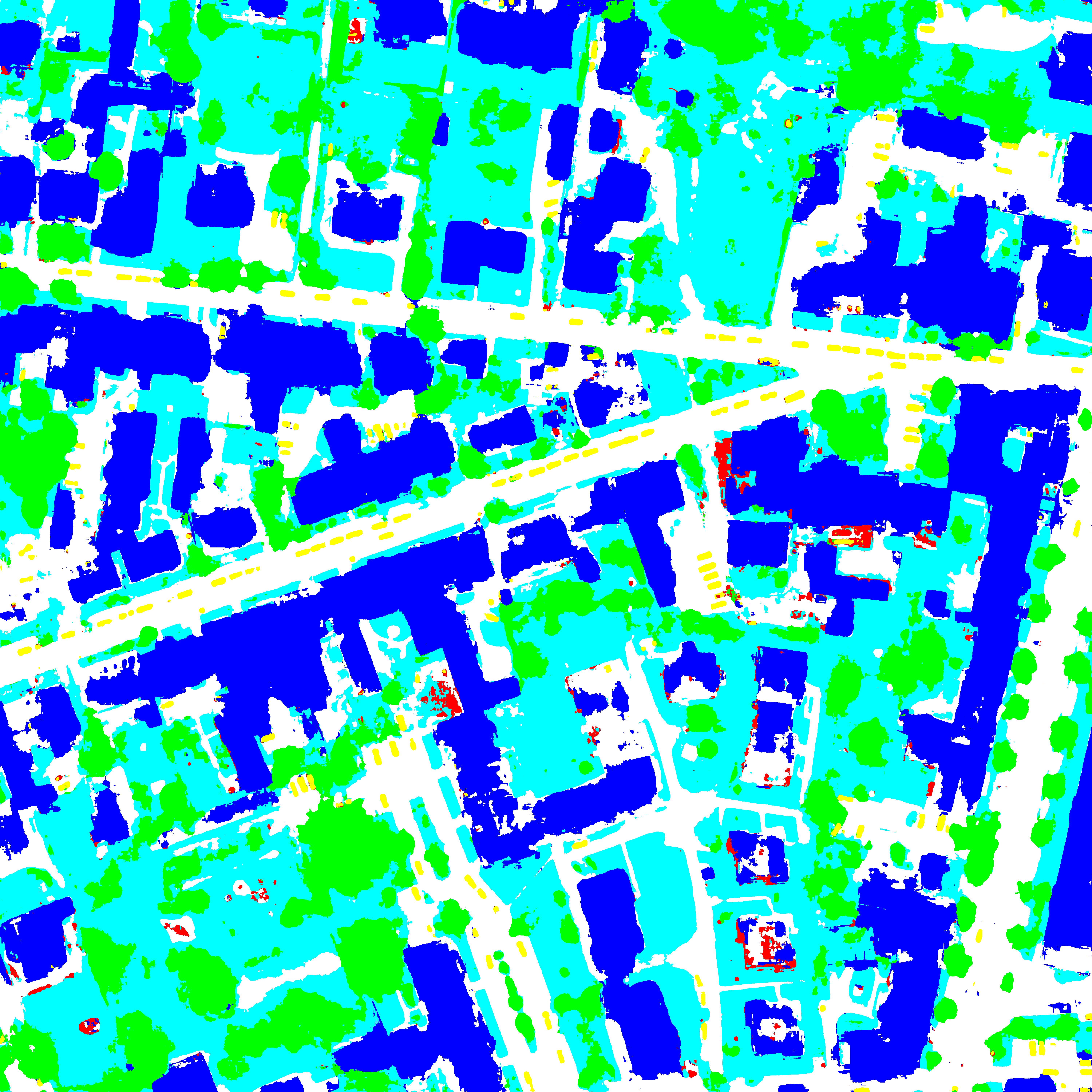}
   \caption{}
\end{subfigure}
\hfill
\begin{subfigure}{.11\textwidth}
   \includegraphics[width=\textwidth]{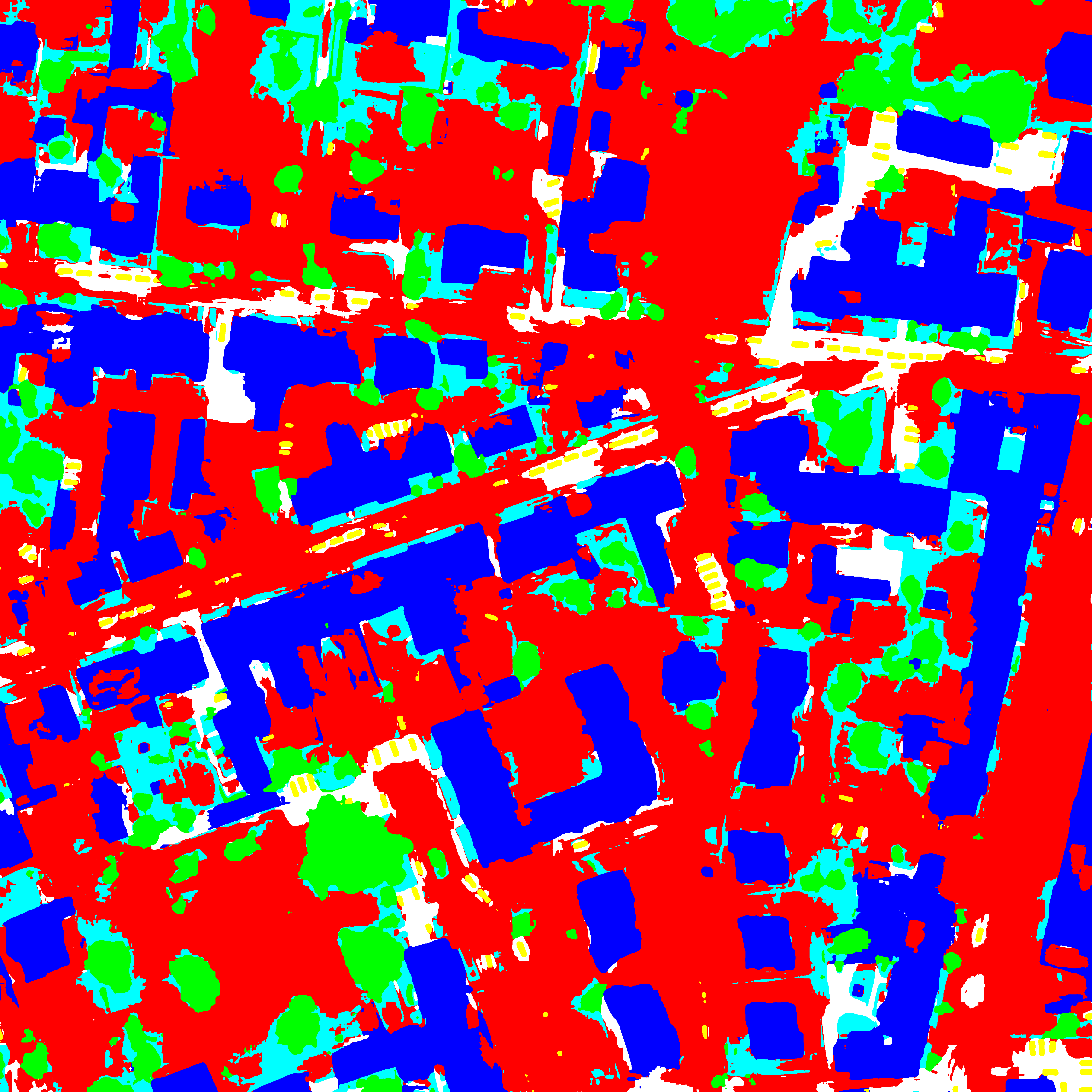}
   \caption{}
\end{subfigure}
\hfill
\begin{subfigure}{.11\textwidth}
   \includegraphics[width=\textwidth]{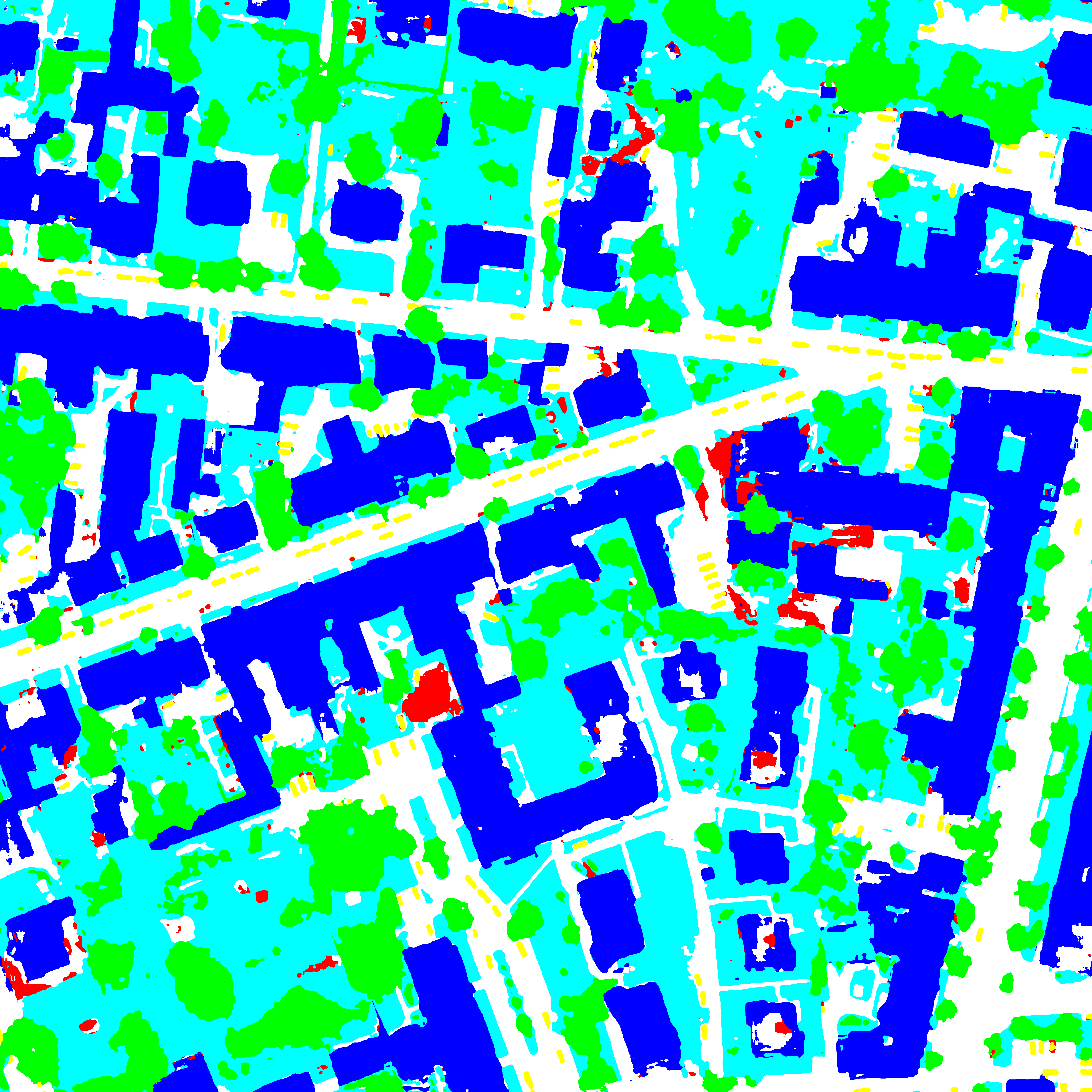}
   \caption{}
\end{subfigure}
\hfill
\begin{subfigure}{.11\textwidth}
   \includegraphics[width=\textwidth]{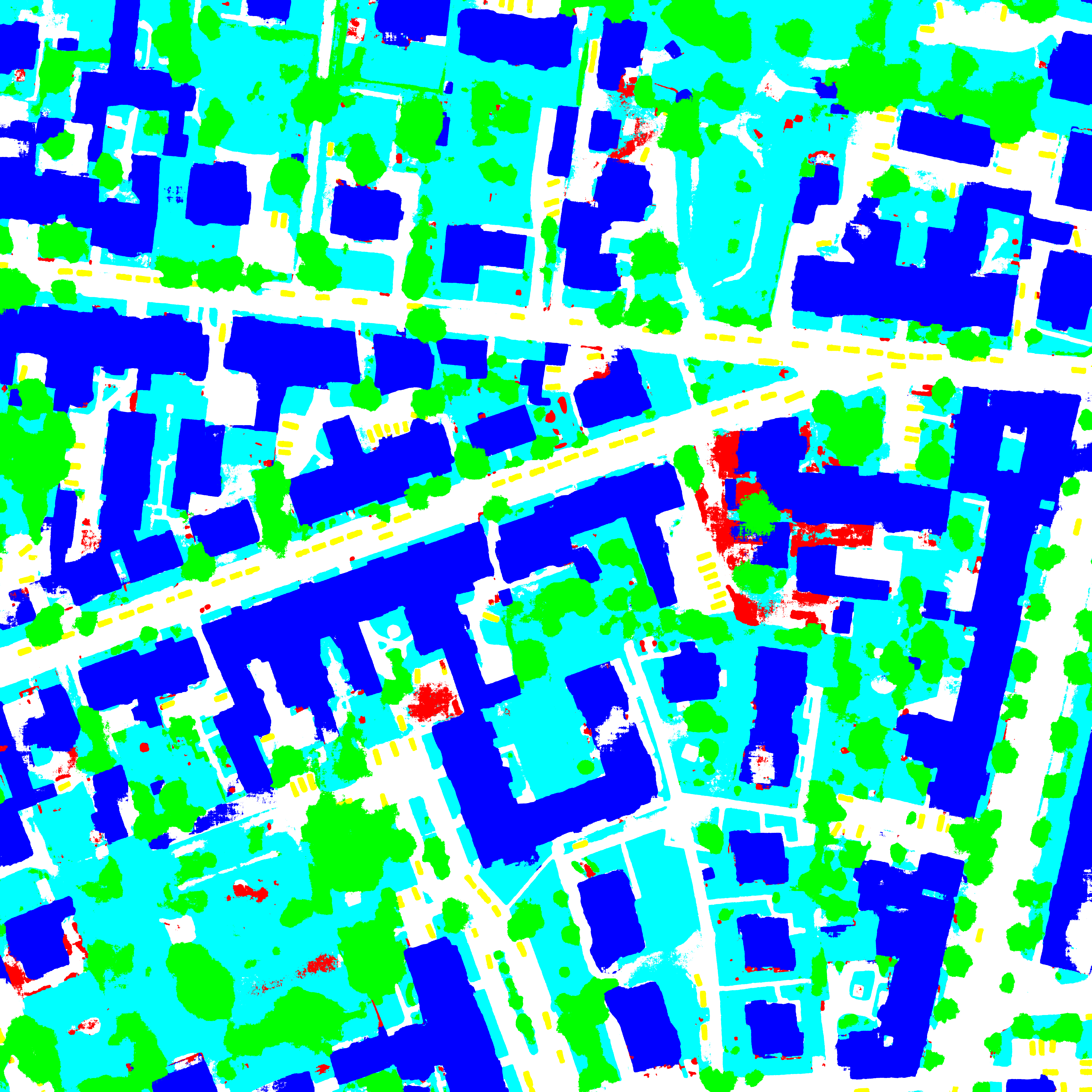}
   \caption{}
\end{subfigure}
\hfill
\begin{subfigure}{.11\textwidth}
   \includegraphics[width=\textwidth]{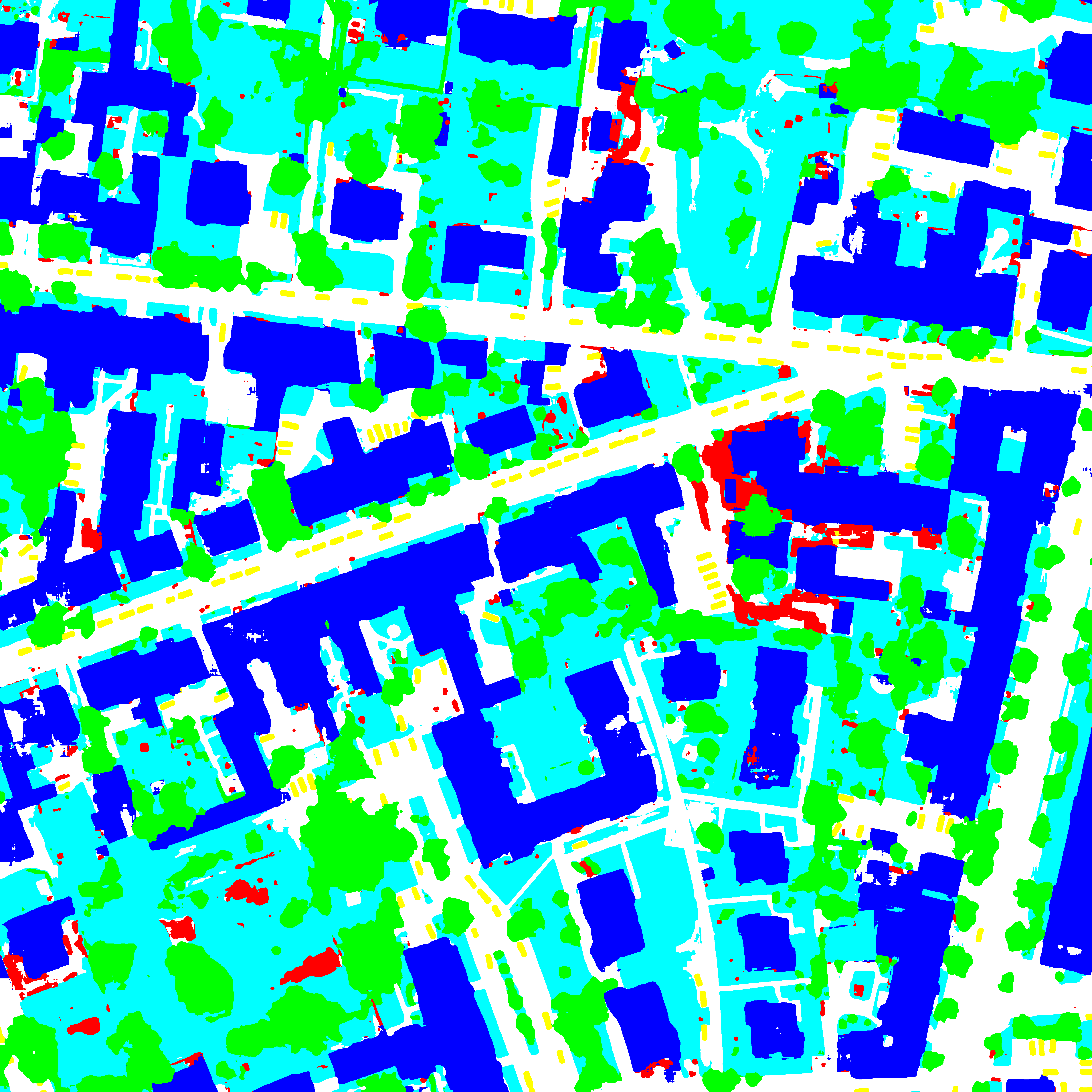}
   \caption{}
\end{subfigure}
\hfill
\begin{subfigure}{.11\textwidth}
   \includegraphics[width=\textwidth]{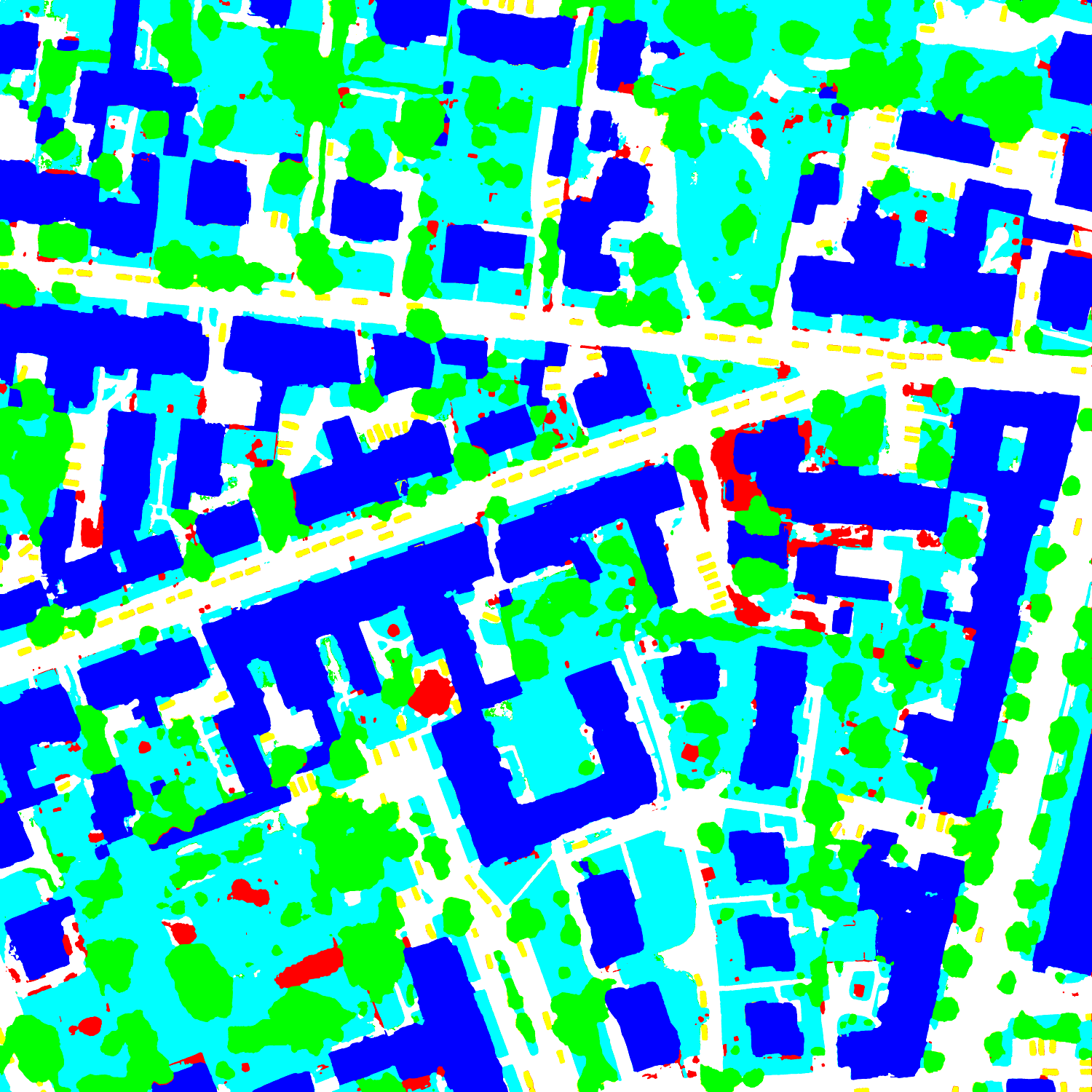}
   \caption{}
\end{subfigure}
\vfill
\begin{subfigure}{.11\textwidth}
   \includegraphics[width=\textwidth]{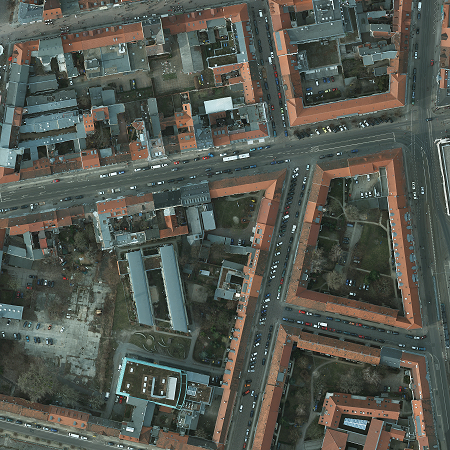}
   \caption{IRRG}
\end{subfigure}
\hfill
\begin{subfigure}{.11\textwidth}
   \includegraphics[width=\textwidth]{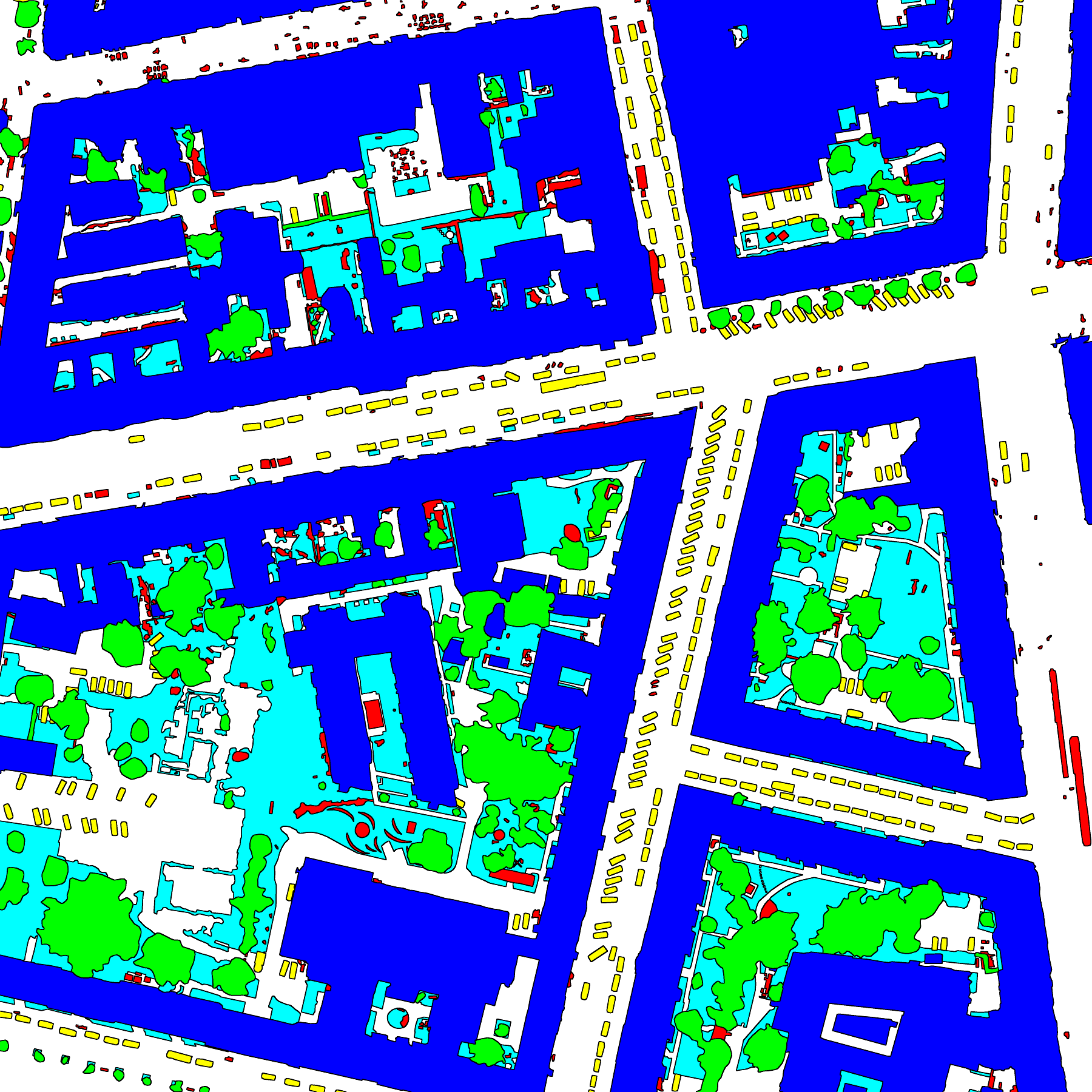}
   \caption{Ground truth}
\end{subfigure}
\hfill
\begin{subfigure}{.11\textwidth}
   \includegraphics[width=\textwidth]{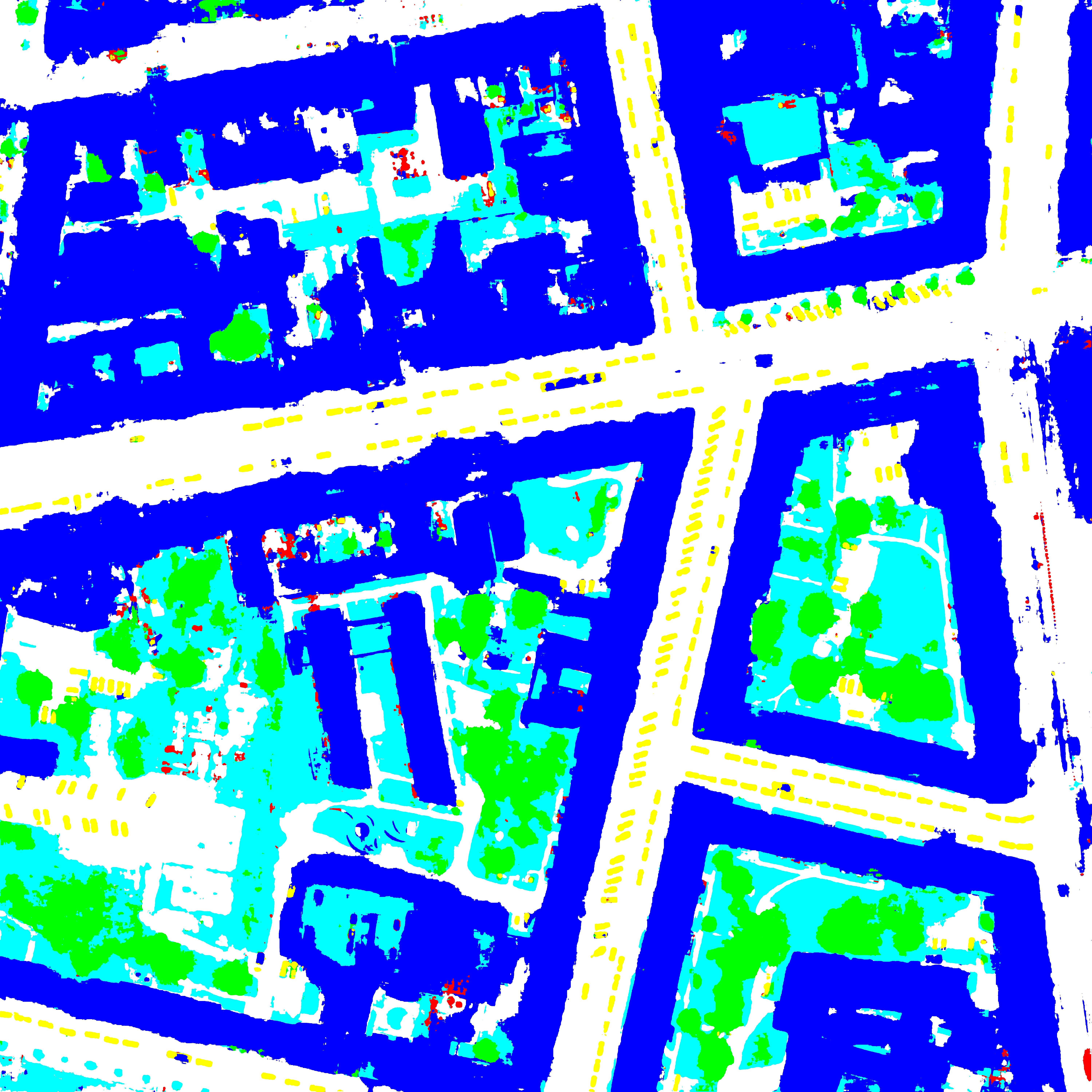}
   \caption{FCN8s}
\end{subfigure}
\hfill
\begin{subfigure}{.11\textwidth}
   \includegraphics[width=\textwidth]{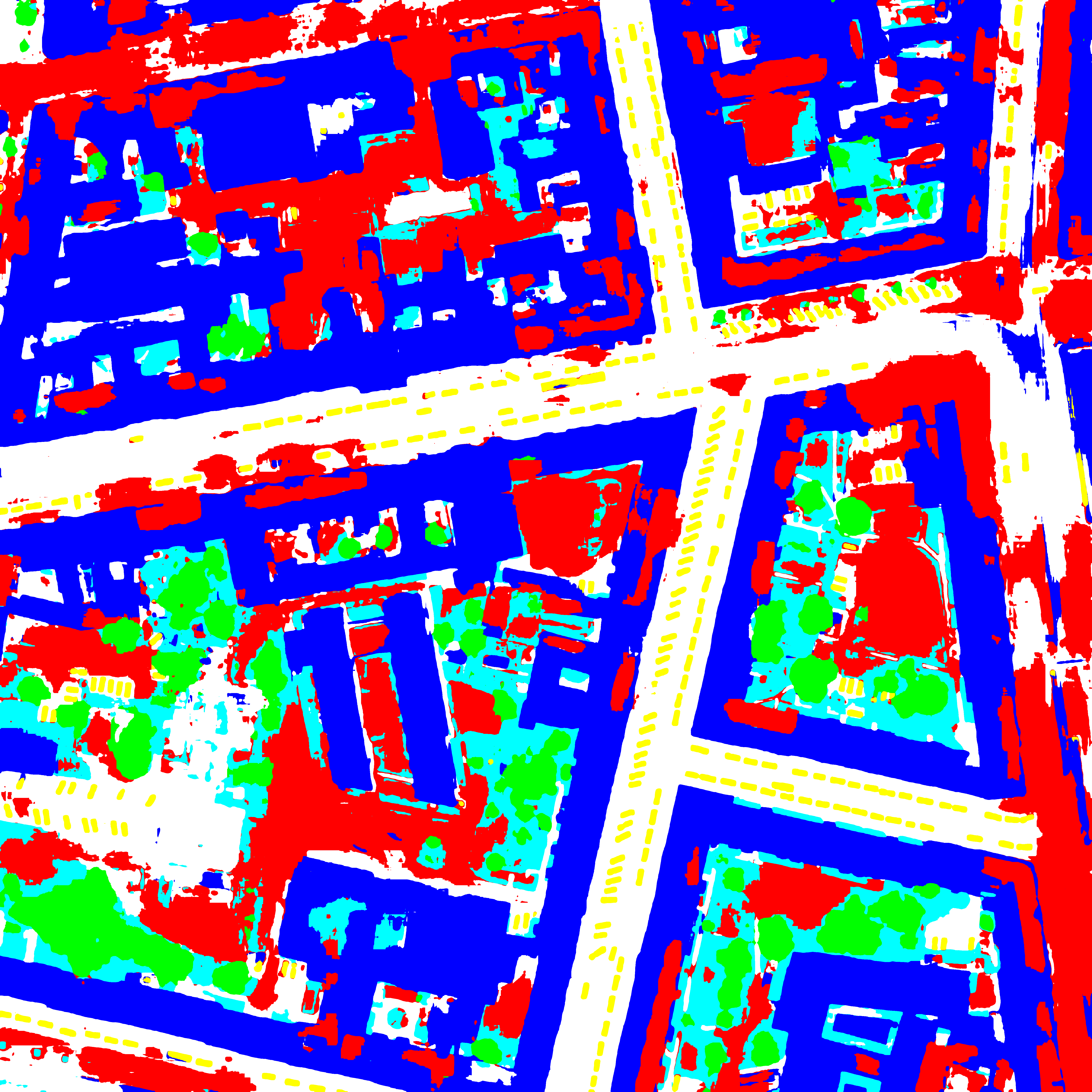}
   \caption{PSPNet}
\end{subfigure}
\hfill
\begin{subfigure}{.11\textwidth}
   \includegraphics[width=\textwidth]{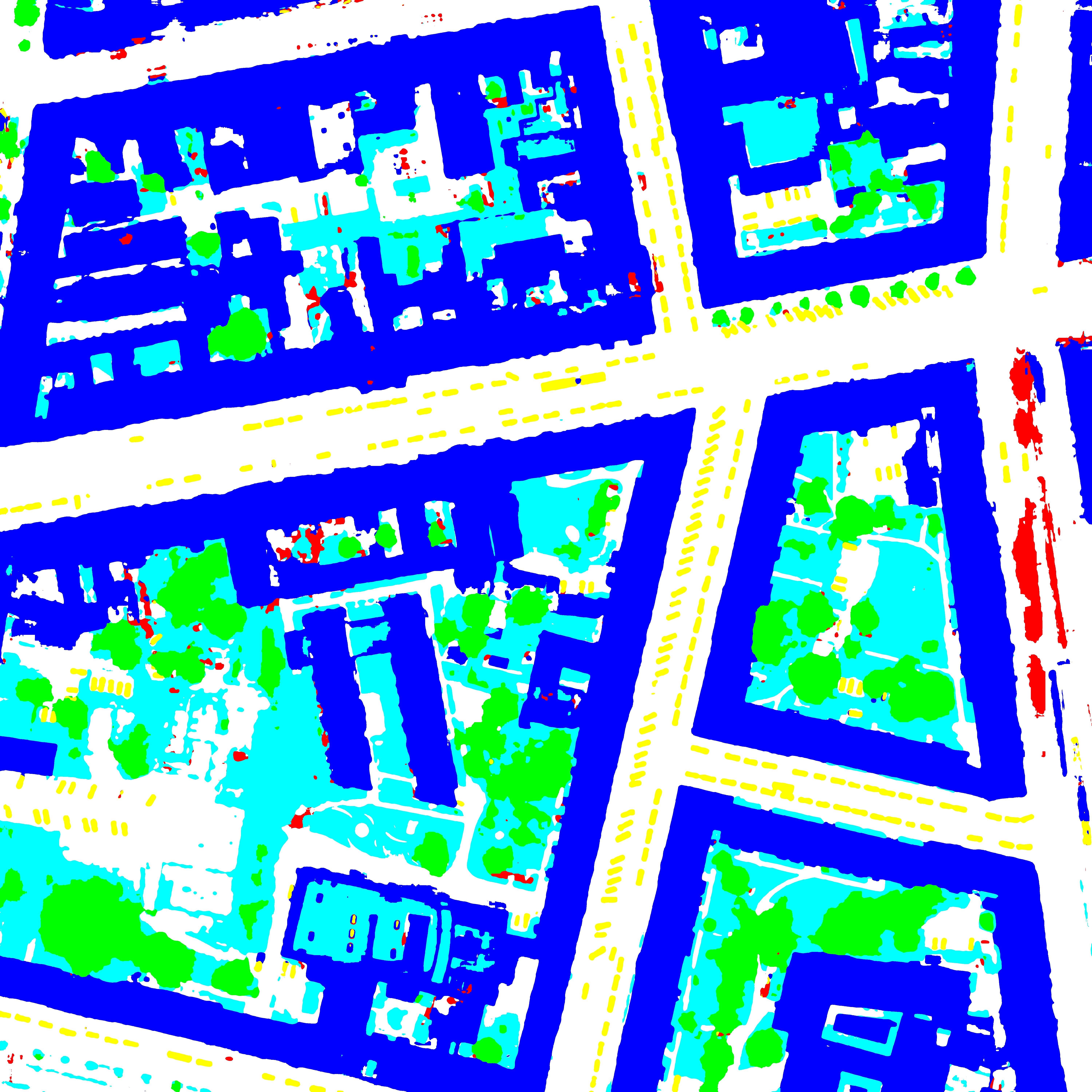}
   \caption{Unet}
\end{subfigure}
\hfill
\begin{subfigure}{.11\textwidth}
   \includegraphics[width=\textwidth]{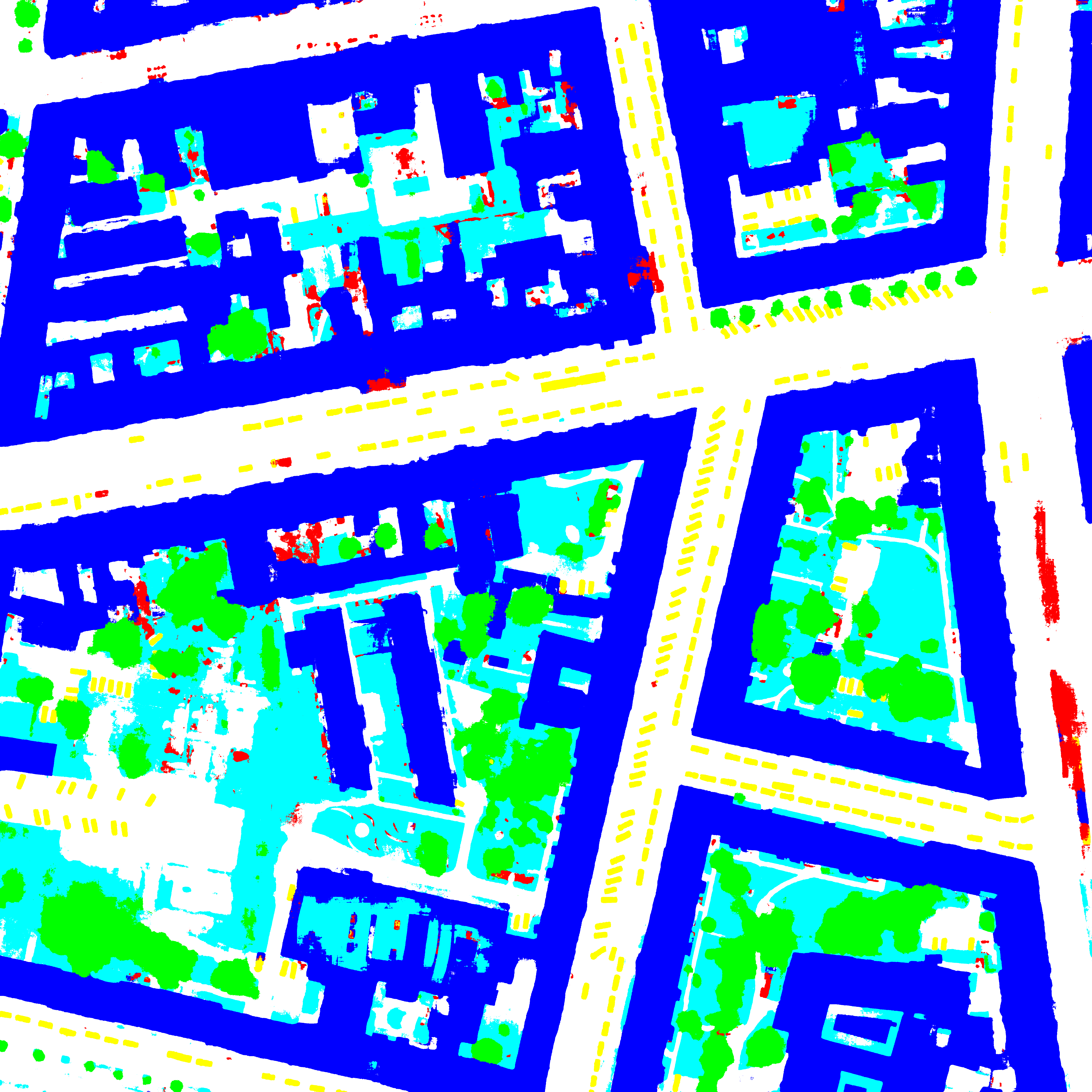}
   \caption{SegNet}
\end{subfigure}
\hfill
\begin{subfigure}{.11\textwidth}
   \includegraphics[width=\textwidth]{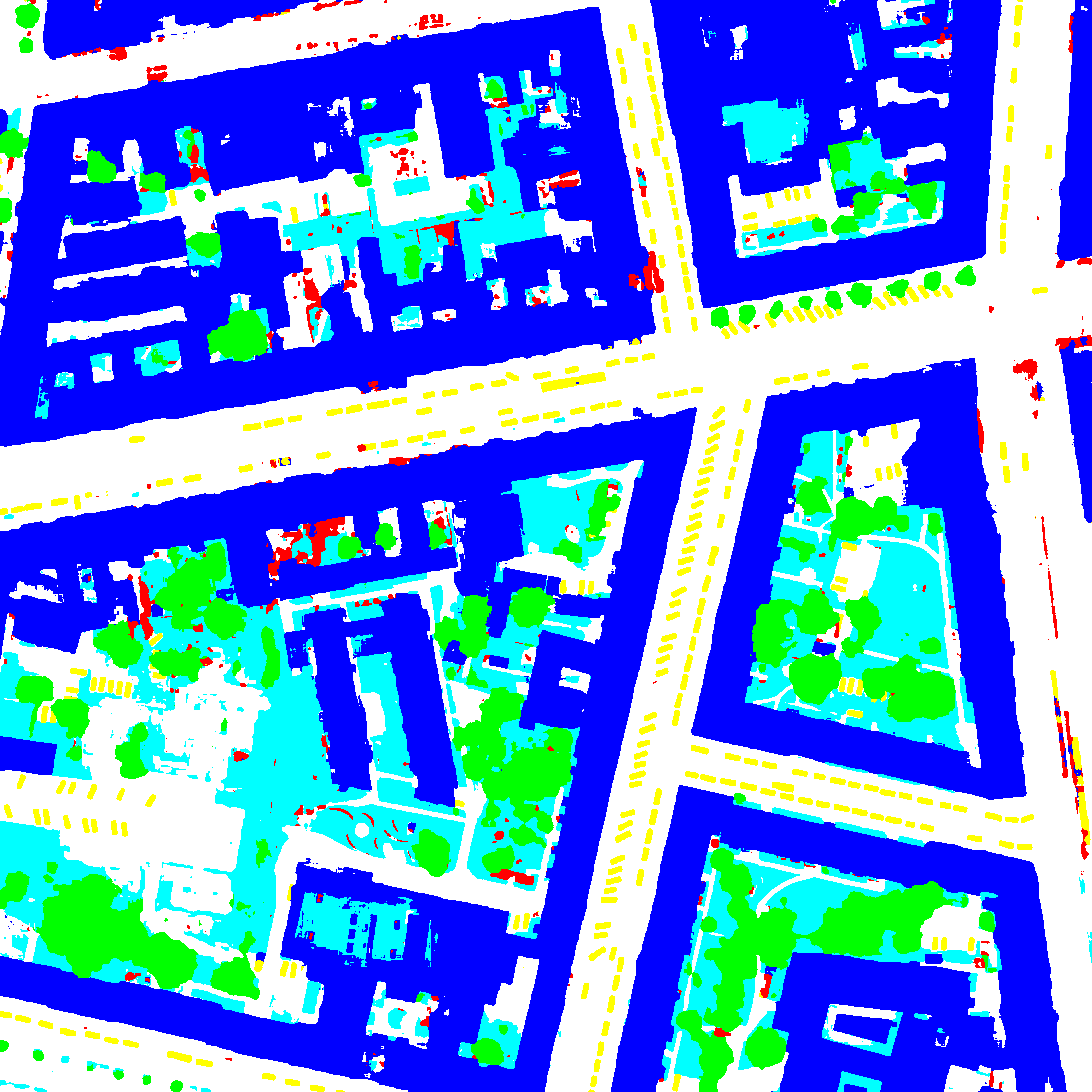}
   \caption{Transformer}
\end{subfigure}
\hfill
\begin{subfigure}{.11\textwidth}
   \includegraphics[width=\textwidth]{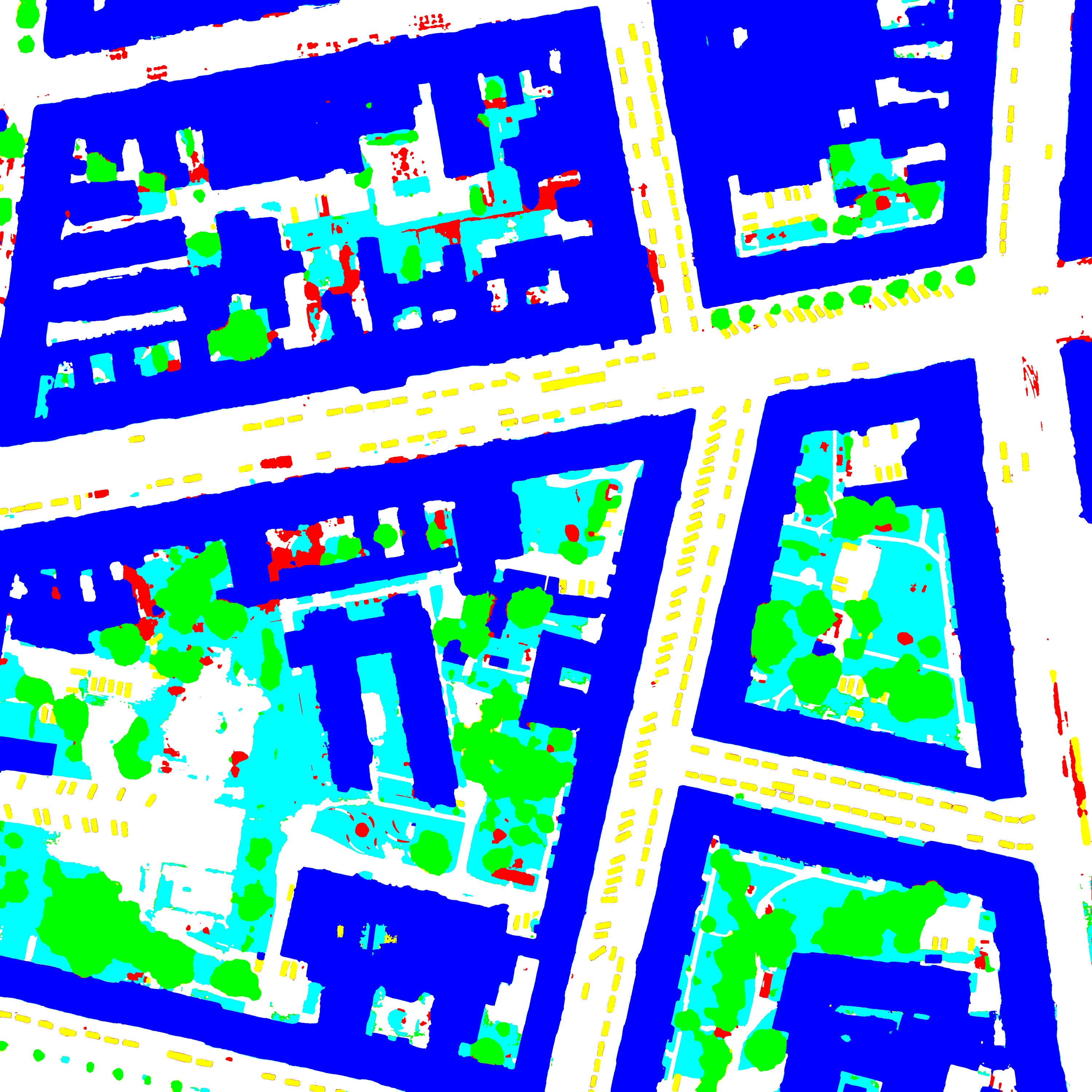}
   \caption{Proposed}
\end{subfigure}

\caption{Performance comparison with other deep learning models using the Potsdam validation dataset. The labels represent the impervious surface in white, buildings in blue, low vegetation in cyan, trees in green, cars in yellow, and clutter/background in red.}
\label{fig: 8}
\end{figure*}

The results obtained for the Vaihingen challenge dataset are presented in Fig.~\ref{fig: 7} and Table~\ref{table:1}, and those for the Potsdam challenge dataset are given in Fig.~\ref{fig: 8} and Table~\ref{table:2}. One can see in Fig.~\ref{fig: 8} and Table~\ref{table:1} that the UNet, FCN-8s, and PSPNet models achieve low-quality results compared to the SegNet, Transformer, and the proposed model. For example, all convolutional models (i.e., UNet, FCN-8s, PSPNet, and SegNet) show low performance in car segmentation; however, the transformer-based models (i.e. transformer alone or our fusion models) classify that category accurately (improvement by more than 3\%). This result demonstrates the superior ability of transformers in representing dynamically changing object classes. The late fusion between the EffUNet and the transformer improved the kappa and overall accuracy by around 1\% compared to the transformer alone or SegNet. We evaluate the model without the semantic tokenizer to see its influence on prediction quality. The results demonstrate its effectiveness, especially in labeling objects from the low vegetation and cars classes, in which we observe a difference of about 3\% in local accuracy (see Table~\ref{table:1}).
\begin{table*}
\renewcommand{\arraystretch}{1.25}
\caption{Performance comparison with other deep learning models on the Vaihingen validation dataset, with the values in bold showing the best-obtained values.}
\centering

\begin{tabular}{lccccccc} 
\hline
\multirow{2}{*}{Model}           & \multicolumn{5}{c}{F1 Score}                                                                                                                         & \multicolumn{1}{l}{\multirow{2}{*}{~
  Kappa}} & \multicolumn{1}{l}{\multirow{2}{*}{Total
  accuracy}}  \\ 
\cline{2-6}
                                 & \multicolumn{1}{l}{Imp. Surf.} & \multicolumn{1}{l}{Buildings} & \multicolumn{1}{l}{Low veg.} & \multicolumn{1}{l}{Trees} & \multicolumn{1}{l}{Cars} & \multicolumn{1}{l}{}                           & \multicolumn{1}{l}{}                                   \\ 
\hline
UNet                             & 90.11                          & 93.97                         & 74.01                        & 88.31                     & 65.23                    & 84.11                                          & 88.25                                                  \\
FCN8s                            & 82.23                          & 87.52                         & 64.51                        & 55.86                     & 44.66                    & 76.01                                          & 82.32                                                  \\
PSPNet                           & 91.13                          & 95.11                         & 78.22                        & 89.81                     & 77.27                    & 86.28                                          & 89.82                                                  \\
SegNet                           & 91.22                          & 95.56                         & 79.16                        & 90.45                     & 83.12                    & 86.96                                          & 90.31                                                  \\
Transformer                      & 91.23                          & 95.69                         & 79.02                        & 90.47                     & 87.83                    & 87.04                                          & 90.37                                                  \\
Proposed model without tokenizer & 91.28                  & 95.13                & 77.75                & 89.88            & 86.36           & 86.40                                 & 89.91                                         \\
Proposed~ model                  & \textbf{91.47}                 & \textbf{96.38}                & \textbf{79.42}               & \textbf{90.89}            & \textbf{88.12}           & \textbf{87.70}                                 & \textbf{90.87}                                         \\
\hline
\end{tabular}
\label{table:1}
\end{table*}

\begin{table*}
\renewcommand{\arraystretch}{1.25}
\caption{Performance comparison with other deep learning models on the Potsdam validation dataset, with the values in bold showing the best-obtained values.}
\centering
\begin{tabular}{lccccccc} 
\hline

\multirow{2}{*}{Model} & \multicolumn{5}{c}{F1 Score}                                                                                                                         & \multicolumn{1}{l}{\multirow{2}{*}{~
  Kappa}} & \multicolumn{1}{l}{\multirow{2}{*}{Total
  accuracy}}  \\ 
\cline{2-6}
                       & \multicolumn{1}{l}{Imp. Surf.} & \multicolumn{1}{l}{Buildings} & \multicolumn{1}{l}{Low veg.} & \multicolumn{1}{l}{Trees} & \multicolumn{1}{l}{Cars} & \multicolumn{1}{l}{}                           & \multicolumn{1}{l}{}                                   \\ 
\hline
UNet                   & 90.00                          & 93.86                         & 85.48                        & 83.31                     & 91.14                    & 84.35                                          & 88.17                                                  \\
FCN8s                  & 84.09                          & 86.12                         & 75.13                        & 72.71                     & 46.95                    & 73.65                                          & 80.22                                                  \\
PSPNet                 & 47.33                          & 81.98                         & 42.15                        & 59.28                     & 89.30                    & 38.85                                          & 47.78                                                  \\
SegNet                 & 91.39                          & 94.81                         & 86.86                        & 83.80                     & \textbf{95.48}           & 85.81                                          & 89.26                                                  \\
Transformer            & \textbf{92.92}                 & 96.24                         & 86.51                        & 87.60                     & 95.45                    & 88.13                                          & 91.04                                                  \\
Proposed model         & 92.86                          & \textbf{97.78}                & \textbf{87.30}               & \textbf{88.72}            & 93.45                    & \textbf{89.31}                                 & \textbf{91.95 }                                        \\
\hline
\end{tabular}
\label{table:2}
\end{table*}

\begin{figure}[hptb]
\begin{subfigure}{.11\textwidth}
   \includegraphics[width=\textwidth]{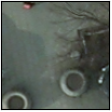}
   \caption{RGB}
\end{subfigure}%
\hfill
\begin{subfigure}{.11\textwidth}
   \includegraphics[width=\textwidth]{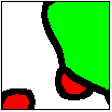}
   \caption{Ground truth}
\end{subfigure}
\hfill
\begin{subfigure}{.11\textwidth}
   \includegraphics[width=\textwidth]{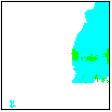}
   \caption{FCN8s}
\end{subfigure}
\hfill
\begin{subfigure}{.11\textwidth}
   \includegraphics[width=\textwidth]{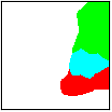}
   \caption{PSPNet}
\end{subfigure}
\vfill
\begin{subfigure}{.11\textwidth}
   \includegraphics[width=\textwidth]{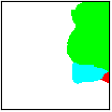}
   \caption{Unet}
\end{subfigure}
\hfill
\begin{subfigure}{.11\textwidth}
   \includegraphics[width=\textwidth]{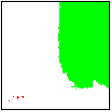}
   \caption{SegNet}
\end{subfigure}
\hfill
\begin{subfigure}{.11\textwidth}
   \includegraphics[width=\textwidth]{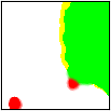}
   \caption{Transformer}
\end{subfigure}
\hfill
\begin{subfigure}{.11\textwidth}
   \includegraphics[width=\textwidth]{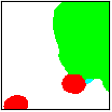}
   \caption{Proposed}
\end{subfigure}
\vfill
\begin{subfigure}{.11\textwidth}
   \includegraphics[width=\textwidth]{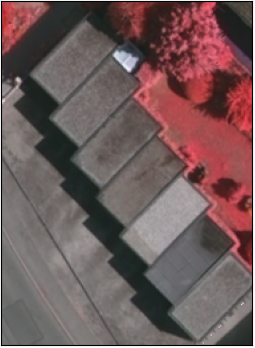}
   \caption{IRRG}
\end{subfigure}%
\hfill
\begin{subfigure}{.11\textwidth}
   \includegraphics[width=\textwidth]{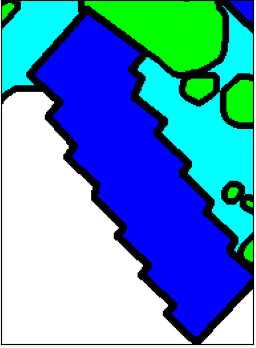}
   \caption{Ground truth}
\end{subfigure}
\hfill
\begin{subfigure}{.11\textwidth}
   \includegraphics[width=\textwidth]{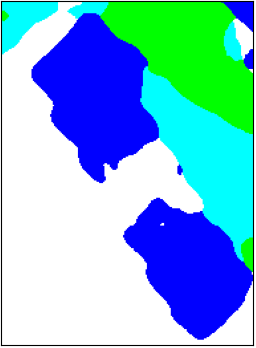}
   \caption{FCN8s}
\end{subfigure}
\hfill
\begin{subfigure}{.11\textwidth}
   \includegraphics[width=\textwidth]{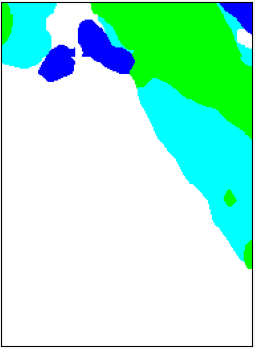}
   \caption{PSPNet}
\end{subfigure}
\vfill
\begin{subfigure}{.11\textwidth}
   \includegraphics[width=\textwidth]{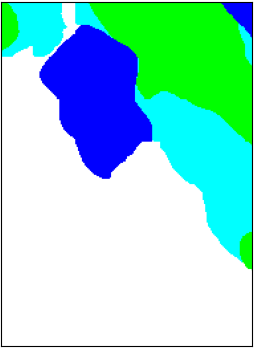}
   \caption{Unet}
\end{subfigure}
\hfill
\begin{subfigure}{.11\textwidth}
   \includegraphics[width=\textwidth]{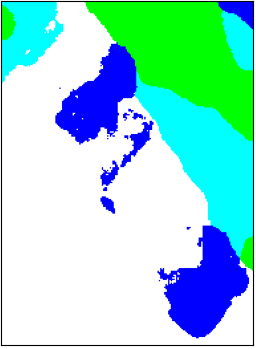}
   \caption{SegNet}
\end{subfigure}
\hfill
\begin{subfigure}{.11\textwidth}
   \includegraphics[width=\textwidth]{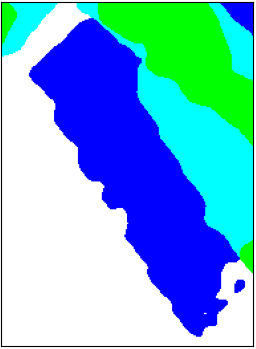}
   \caption{Transformer}
\end{subfigure}
\hfill
\begin{subfigure}{.11\textwidth}
   \includegraphics[width=\textwidth]{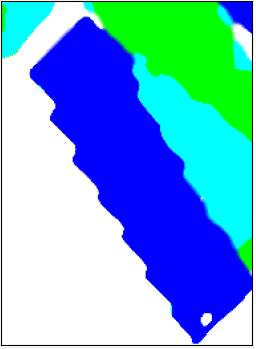}
   \caption{Proposed}
\end{subfigure}

\caption{Model performance when dealing with unclear objects. The labels represent impervious surfaces in white, buildings in blue, low vegetation in cyan, trees in green, cars in yellow, clutter/background in red, and rectangles indicate confused objects}
\label{fig: 9}
\end{figure}

The results of the Potsdam dataset show that the transformer model alone improved the results by more than 2\% in kappa and by around 1\% in total accuracy compared to the SegNet model, which achieved the closest results. The fusion adds 1\% to the overall accuracy. The transformer alone and fused with EffU-Net each improve the tree's classification quality by around 4\% compared to the conventional models, such as SegNet or UNet. To demonstrate the ability of the proposed model when dealing with unclear objects, we took a sample that contained such objects (Fig.~\ref{fig: 9}). We can see that the proposed model predicts these objects well, compared to other models.

\begin{table*}[h!]
\renewcommand{\arraystretch}{1.25}
\centering
\caption{Performance comparison with other deep learning models on the Vaihingen test dataset, with the values in bold showing the best-obtained values.}
\begin{tabular}{lcccccc} 
\hline
\multicolumn{1}{c}{\multirow{2}{*}{Method}} & \multicolumn{5}{c}{F1 Score}  & \multicolumn{1}{l}{\multirow{2}{*}{Overall Accuracy}} \\ 
\cline{2-6}
\multicolumn{1}{c}{}  & \multicolumn{1}{l}{Imp. Surface} & \multicolumn{1}{l}{Building} & \multicolumn{1}{l}{Low Veg.} & \multicolumn{1}{l}{Tree} & \multicolumn{1}{l}{Car} & \multicolumn{1}{l}{}                                     \\ 
\hline
UFMG\_4 \cite{7725499} 
                        & 91.1  & 94.5    & 82.9   & 88.8    & 81.3  & 89.4        \\
CASIA2 \cite{LIU201878}          
                        & 93.2  & 96.0    & 84.7   & 89.9    & 86.7  & 91.1         \\
DLR\_9 \cite{MARMANIS2018158}    
                       & 92.4  & 95.2    & 83.9   & 89.9    & 81.2  & 90.3          \\
HUSTW3 \cite{SUN2019297}                                 
                      & 92.1   & 95.3    & 85.6   & 90.5    & 78.3  & 90.7          \\
ONE\_7 \cite{AUDEBERT201820}
                      & 91.0   & 94.5    & 84.4   & 89.9    & 77.8  & 89.8          \\
ADL\_3 \cite{Peng_2017_CVPR}                                 
                     & 89.5    & 93.2    & 82.3   & 88.2   & 63.3   & 88.0           \\
DST\_2 \cite{7301381}                                
                     & 90.5    & 93.7                         & 83.4                           & 89.2                     & 72.6                    & 89.1                                                     \\
GSN3 \cite{https://doi.org/10.48550/arxiv.1606.02585}                                   & 92.2                               & 95.1                         & 83.7                           & 89.9                     & 82.4                    & 90.3                                                     \\
UZ\_1 \cite{rs9050446}                                  & 89.2                               & 92.5                         & 81.6                           & 86.9                     & 57.3                    & 87.3                                                     \\
RIT\_7 \cite{10.1007/978-3-642-75988-8_28}                                 & 91.7                               & 95.2                         & 83.5                           & 89.2                     & 82.8                    & 89.9                                                     \\
LANet \cite{9102424}                                  & 92.4                               & 94.9                         & 82.9                           & 88.9                     & 81.3                    & 89.9                                                     \\
Swin-B-CNN+BD \cite{9686732}                          & 92.2                               & 95.3                         & 83.6                           & 89.6                     & 86.9                    & 90.4                                                     \\
AREANs-ResNeSt \cite{doi:10.1080/2150704X.2021.1910362}                         & 93.2                               & 96.1                         & 84.8                           & 90.2                     & 90.5                    & 91.3                                                     \\
MF-DFNet \cite{doi:10.1080/01431161.2021.2018147}                              & 88.8                               & 93.1                         & 78.4                           & 84.0                     & 77.5                    & 86.2                                                     \\
DGCR \cite{9690154}                                   & 92.9                               & 95.8                         & 84.7                           & 90.1                     & 86.5                    & 91.1                                                     \\
Swin-S \cite{9681903}                                 & 93.6                               & 96.2                         & 85.8                           & 90.4                     & 87.6                    & 91.6                                                     \\
CEGFNet \cite{9538389}                                & -                                  & -                            & -                              & -                        & -                       & 89.3                                                     \\
G2GNet \cite{9519842}                                 & 92.1                               & 94.8                         & 83.8                           & 89.6                     & 85.4                    & 90.2                                                     \\
AFNet \cite{9258402}                                  & 93.4                               & 95.9                         & 86.0                           & \textbf{90.7}            & 87.2                    & 91.6                                                     \\
SBANet \cite{9345482}                                 & 94.4                               & 92.9                         & 83.4                           & 89.6                     & 91.4                    & 90.5                                                     \\
SVL \cite{rs10091429}                                   & 86.6                               & 91.0                         & 77.0                           & 85.0                     & 55.6                    & 84.8                                                     \\
RIT\_L \cite{10.1117/12.2243169}                                 & 89.6                               & 92.2                         & 81.6                           & 88.6\textbf{}            & 76.0                    & 87.8                                                     \\

Proposed model (IRRG)                       & 93.5                               & 95.7                         & 84.9                           & 89.9                     & 87.8                    & 91.5                                                     \\
Without decoder (IRRG +DSM)                 & 93.5                               & 96.2                         & 84.8                           & 90.4                     & 87.1                    & 91.5                                                     \\
\textbf{Proposed model (IRRG +DSM)}         & \textbf{94.7}                      & \textbf{98.0}                & \textbf{89.4}                  & 90.4                     & \textbf{96.0}           & \textbf{91.8}   \\
\hline
\end{tabular}
\label{table:3}
\end{table*}

\begin{table*}[h!]
\renewcommand{\arraystretch}{1.25}
\centering
\caption{Performance comparison with other deep learning models on the Potsdam test dataset, with the values in bold showing the best-obtained values.}
\begin{tabular}{lcccccc} 
\hline
\multicolumn{1}{c}{\multirow{2}{*}{Method}} & \multicolumn{5}{c}{F1 Score}  & \multicolumn{1}{l}{\multirow{2}{*}{Overall Accuracy}} \\ 
\cline{2-6}
\multicolumn{1}{c}{}  & \multicolumn{1}{l}{Imp. Surface} & \multicolumn{1}{l}{Building} & \multicolumn{1}{l}{Low Veg.} & \multicolumn{1}{l}{Tree} & \multicolumn{1}{l}{Car} & \multicolumn{1}{l}{}                                     \\ 
\hline
CASIA2 \cite{LIU201878}    
&93.3                     & 97.0                 & 87.7                   & 88.4             & 96.2            & 91.1       \\

HUSTW3 \cite{SUN2019297}  
& 93.6                   &97.6        & 88.5                   & 88.8             & 94.6         & 91.6                                             \\
DST \cite{7301381}                          & 92.5                     & 96.4                 & 86.7                   & 88.0            & 94.7            & 90.3  \\        
UZ \cite{rs9050446}                            & 89.3                     & 95.4          & 81.8            & 80.5     & 86.5      & 85.8 \\                        
RIT \cite{8727958}                  & 92.6             & 97.0        & 86.9           & 87.4       & 95.2      & 90.3 \\                                     
LANet \cite{9102424}                         & 93.1            & 97.2            & 87.3         & 88.0        & 94.2       & 90.8   \\
Swin-B-CNN+BD \cite{9686732}      & 93.6       & 96.7          & 88.0          & 88.0      & 96.3        & 91.0 \\                                                  
AREANs-ResNeSt \cite{doi:10.1080/2150704X.2021.1910362}                
                 & 94.3    & 97.4  & 88.5       & 89.6   & 97.0           & 91.9                 \\               
DGCR \cite{9690154}             & 94.1                   &97.3                 & 88.3         & 88.9   & 93.0           & 91.8                                             \\                        
Swin-S \cite{9681903}             & 94.2        & 97.6         & 88.6             & 89.6         & 96.3       & 92.0                                          \\    
CEGFNet \cite{9538389}                       & {-}                        & {-}                    & {-}                      & {-}                & {-}               & 85.2  \\                                       
G2GNet \cite{9519842}                       & 93.0                   & 96.5              & 86.3                 & 88.2             & \textbf{95.8}   & 90.3                                             \\
AFNet \cite{9258402}                         &94.2                  & 97.2                & 89.2                   & 89.4           & 95.1            & 92.2                                           \\
SBANet \cite{9345482}                       & 93.8                   & \textbf{98.0}      & 89.0                 & 89.5       & 94.7          & 92.8                                           \\
SVL \cite{rs10091429}                 & 83.5                   & 91.7               & 72.2                 & 63.2            & 62.2   & 77.8                                             \\
RIT\_L \cite{10.1117/12.2243169}                         & 91.2                     & 94.6                 & 85.1                  & 85.1             & 92.8           & 88.4                                             \\

Proposed model (RGB)            & 93.5                    & 97.2              & 86.9                  & 88.5             & 96.2          & 91.8                                           \\
Without~decoder~(RGB +DSM)       & 94.0                     & \textbf{98.0}        & 89.2                   & 90.0             & 95.8            & 92.5                                             \\
\textbf{Proposed model (RGB +DSM)} & \textbf{94.8}           & \textbf{98.0}        & \textbf{89.5}          & \textbf{90.5}    & 94.6           & \textbf{92.9}                           \\
\hline
\end{tabular}
\label{table:4}
\end{table*}
\textbf{ISPRS benchmark dataset:} The proposed model was tested using the testing dataset available on the website of the ISPRS challenge \cite{labeling_2016}. The performance of the model was compared with the following participants in the challenge:

$1)~ SegNet+DSM+NDSM ~(ONE\_7)$ \cite{AUDEBERT201820}:The authors used the late fusion of two trained SegNets using the IRRG image and the composite image that contained NDVI, DSM, and NDSM.

$2)	~Self-cascaded+ResNet (CASIA2)$ \cite{LIU201878}: A single self-cascaded network with the encoder based on a variant of a 101-layer ResNet [38]. The authors used only the 3-band IRRG images to predict the segments, which makes their model computationally more tractable. 

$3)	~CNN+HCF+CRF ~(ADL\_3)$ \cite{Peng_2017_CVPR} : The model used a CNN to extract the image features to produce per-pixel category probabilities; then, a conditional random field (CRF) was applied as a post-processing step to find the predicted labels.

$4)	~FCN+SegNet+NDSM ~(DLR\_9)$ \cite{MARMANIS2018158}: The authors used an ensemble model that contained an FCN and SegNet based on the IRRG and NDSM images to find the best label segmentation.

$5)	~FCN+DSM+RF+CRF (DST\_2)$ \cite{7301381}: IRRG and DSM data were used as inputs to the combined model based on the FCN trained with no downsampling and random forest to find the output features; next, the CRF is used as a post-processing step.

$6)	~Gated~segmentation~network~GSN3$ \cite{https://doi.org/10.48550/arxiv.1606.02585}: A gated segmentation network was proposed. ResNet{-}101 was used as the feature extractor in the encoder portion, and the entropy control module was used for feature fusion in the decoder. A residual convolution module (RCM) was employed as the basic processing unit.

$7)	~CNN+NDSM+Deconvolution~(UZ\_1)$ \cite{rs9050446}: The model is comprised of a CNN that has been trained to learn a series of downsampling (a regular CNN) and a sequence of nonlinear upsampling blocks using deconvolutions back to the original input size.

$8)	~Dilated~Convnet~ UFMG\_4$ \cite{7725499}: The authors proposed a series of dilated convolutions \cite{8727958}. The primary concept is to train a dilated network with different patch sizes to collect multi-context features from diverse contexts.

$9)	~SegNet+FCN~ (RIT\_7)$ \cite{10.1007/978-3-642-75988-8_28}: In their model, SegNet was fused with an FCN for pixel-wise semantic classification.

$10) ~LANet$ \cite{9102424}: The authors proposed the local attention network to improve the semantic segmentation of RSIs by enhancing the scene-related representation in both encoding and decoding phases.

$11) ~Swin-B-CNN+BD$ \cite{9686732}: The swin transformers and CNN were used as encoder and decoder for remote sensing segmentation. The CNN is applied to recover the size of the feature maps and acquire the semantic segmentation results backbone.

$12) ~ResNeSt$ \cite{doi:10.1080/2150704X.2021.1910362}: Attention-Residual block- Embedded Adversarial Network was investigated to learn local-to-global contextual information through semantic and position information improved collection.

$13)~MF-DFNet$ \cite{doi:10.1080/01431161.2021.2018147}: A multiscale feature and discriminative feature network was proposed to resolve the issues of intraclass inconsistency and the difficulty in locating and identifying the target.

$14)~DGCR$ \cite{9690154}: A dynamic graph contextual reasoning module over global reasoning networks was presented for capturing long-range dependencies in feature representations.

$15)~Swin-S$ \cite{9681903}: The Swin-transformer and the densely connected feature aggregation module were used as encoder- decoder, respectively, to improve the remote sensing semantic segmentation accuracy. 

$16)~CEGFNet$ \cite{9538389}: An end-to-end common extraction and gate fusion network was proposed to solve the problem of misclassification of small objects. 

$17)~G2GNet$ \cite{9519842}: To calibrate the RGB responses for improved feature representation, the informative features from the RGB and auxiliary data were adaptively gathered using a self-adaptive attention mechanism. 

$18)~AFNet$ \cite{9258402}:  The multiscale and multilevel maps based CNN were combined for remote sensing semantic segmentation. 

$19)~SBANet$ \cite{9345482}:  To extract full and crisp borders from complicated very-high-resolution remote sensing images, a semantic boundary awareness network was developed.

$20)~ HUSTW3$ \cite{rs10091429}: The authors developed a residual architecture for encoder-decoder models to address the issues of inadequate learning and receptive field imbalance faced by encoder-decoder models 

Table.~\ref{table:3} shows that the proposed fusion model outperforms existing methods on the Vaihingen dataset, with a testing accuracy of 91.5\% when using IRRG and 91.8\% when combining IRRG and DSM. 
The proposed approach achieved the highest accuracy (91.8\%) on the Potsdam dataset when using only RGB images, and 92.9 \% when combining RGB with DSM (Table.~\ref{table:4}). We can see that elevation data increased the accuracy only slightly, which makes the RGB image alone preferable for use in systems with computational constraints.
\section{Conclusion}
In this paper, we proposed a novel fusion deep learning model for investigating the semantic labeling of multi-modal ultra-high-resolution urban remote sensing data. We showed that the fusion of deep transformers and conventional neural networks (i.e., the U-Net model) is an effective method for recognizing the relationships between objects and scenes, leading to consistent labeling outcomes for complex urban objects.

Extensive experiments on two publicly available challenging datasets demonstrate the proposed model's efficacy and efficiency. Proposed model was shown to be more consistent and yields more accurate labeling outcomes than existing frameworks.

\section*{Acknowledgment}

The authors would like to thank Dr. Markus Gerke, Technische Universität Braunschweig, for providing the normalized DSM data. This work has been funded by the ICT Fund, Telecommunications Regulatory Authority (TRA), PO. Box: 26662 Abu Dhabi, United Arab Emirates.




\bibliographystyle{IEEEtran}
\bibliography{bare_jrnl.bib}

\begin{thebibliography}{10}
\providecommand{\url}[1]{#1}
\csname url@samestyle\endcsname
\providecommand{\newblock}{\relax}
\providecommand{\bibinfo}[2]{#2}
\providecommand{\BIBentrySTDinterwordspacing}{\spaceskip=0pt\relax}
\providecommand{\BIBentryALTinterwordstretchfactor}{4}
\providecommand{\BIBentryALTinterwordspacing}{\spaceskip=\fontdimen2\font plus
\BIBentryALTinterwordstretchfactor\fontdimen3\font minus
  \fontdimen4\font\relax}
\providecommand{\BIBforeignlanguage}[2]{{%
\expandafter\ifx\csname l@#1\endcsname\relax
\typeout{** WARNING: IEEEtran.bst: No hyphenation pattern has been}%
\typeout{** loaded for the language `#1'. Using the pattern for}%
\typeout{** the default language instead.}%
\else
\language=\csname l@#1\endcsname
\fi
#2}}
\providecommand{\BIBdecl}{\relax}
\BIBdecl

\bibitem{WANG2022104969}
Z.~Wang, J.~Wang, K.~Yang, L.~Wang, F.~Su, and X.~Chen, ``Semantic segmentation
  of high-resolution remote sensing images based on a class feature attention
  mechanism fused with deeplabv3+,'' \emph{Computers \& Geosciences}, vol. 158,
  p. 104969, 2022.

\bibitem{MA2019166}
L.~Ma, Y.~Liu, X.~Zhang, Y.~Ye, G.~Yin, and B.~A. Johnson, ``Deep learning in
  remote sensing applications: A meta-analysis and review,'' \emph{ISPRS
  Journal of Photogrammetry and Remote Sensing}, vol. 152, pp. 166--177, 2019.

\bibitem{MI2020140}
L.~Mi and Z.~Chen, ``Superpixel-enhanced deep neural forest for remote sensing
  image semantic segmentation,'' \emph{ISPRS Journal of Photogrammetry and
  Remote Sensing}, vol. 159, pp. 140--152, 2020.

\bibitem{YUAN2021114417}
X.~Yuan, J.~Shi, and L.~Gu, ``A review of deep learning methods for semantic
  segmentation of remote sensing imagery,'' \emph{Expert Systems with
  Applications}, vol. 169, p. 114417, 2021.

\bibitem{9081937}
H.~Li, K.~Qiu, L.~Chen, X.~Mei, L.~Hong, and C.~Tao, ``Scattnet: Semantic
  segmentation network with spatial and channel attention mechanism for
  high-resolution remote sensing images,'' \emph{IEEE Geoscience and Remote
  Sensing Letters}, vol.~18, no.~5, pp. 905--909, 2021.

\bibitem{7725499}
M.~Volpi and D.~Tuia, ``Dense semantic labeling of subdecimeter resolution
  images with convolutional neural networks,'' \emph{IEEE Transactions on
  Geoscience and Remote Sensing}, vol.~55, no.~2, pp. 881--893, 2017.

\bibitem{7890382}
W.~Zhao, S.~Du, and W.~J. Emery, ``Object-based convolutional neural network
  for high-resolution imagery classification,'' \emph{IEEE Journal of Selected
  Topics in Applied Earth Observations and Remote Sensing}, vol.~10, no.~7, pp.
  3386--3396, 2017.

\bibitem{ZHAO201748}
W.~Zhao, S.~Du, Q.~Wang, and W.~J. Emery, ``Contextually guided
  very-high-resolution imagery classification with semantic segments,''
  \emph{ISPRS Journal of Photogrammetry and Remote Sensing}, vol. 132, pp.
  48--60, 2017.

\bibitem{MARCOS201896}
D.~Marcos, M.~Volpi, B.~Kellenberger, and D.~Tuia, ``Land cover mapping at very
  high resolution with rotation equivariant cnns: Towards small yet accurate
  models,'' \emph{ISPRS Journal of Photogrammetry and Remote Sensing}, vol.
  145, pp. 96--107, 2018, deep Learning RS Data.

\bibitem{8388225}
J.~R. Bergado, C.~Persello, and A.~Stein, ``Recurrent multiresolution
  convolutional networks for vhr image classification,'' \emph{IEEE
  Transactions on Geoscience and Remote Sensing}, vol.~56, no.~11, pp.
  6361--6374, 2018.

\bibitem{LIU201878}
\BIBentryALTinterwordspacing
Y.~Liu, B.~Fan, L.~Wang, J.~Bai, S.~Xiang, and C.~Pan, ``Semantic labeling in
  very high resolution images via a self-cascaded convolutional neural
  network,'' \emph{ISPRS Journal of Photogrammetry and Remote Sensing}, vol.
  145, pp. 78--95, 2018, deep Learning RS Data. [Online]. Available:
  \url{https://www.sciencedirect.com/science/article/pii/S0924271617303854}
\BIBentrySTDinterwordspacing

\bibitem{MARMANIS2018158}
\BIBentryALTinterwordspacing
D.~Marmanis, K.~Schindler, J.~Wegner, S.~Galliani, M.~Datcu, and U.~Stilla,
  ``Classification with an edge: Improving semantic image segmentation with
  boundary detection,'' \emph{ISPRS Journal of Photogrammetry and Remote
  Sensing}, vol. 135, pp. 158--172, 2018. [Online]. Available:
  \url{https://www.sciencedirect.com/science/article/pii/S092427161630572X}
\BIBentrySTDinterwordspacing

\bibitem{SUN2019297}
\BIBentryALTinterwordspacing
Y.~Sun, Y.~Tian, and Y.~Xu, ``Problems of encoder-decoder frameworks for
  high-resolution remote sensing image segmentation: Structural stereotype and
  insufficient learning,'' \emph{Neurocomputing}, vol. 330, pp. 297--304, 2019.
  [Online]. Available:
  \url{https://www.sciencedirect.com/science/article/pii/S0925231218313821}
\BIBentrySTDinterwordspacing

\bibitem{7729406}
Z.~Zhong, J.~Li, W.~Cui, and H.~Jiang, ``Fully convolutional networks for
  building and road extraction: Preliminary results,'' in \emph{2016 IEEE
  International Geoscience and Remote Sensing Symposium (IGARSS)}, 2016, pp.
  1591--1594.

\bibitem{AUDEBERT201820}
N.~Audebert, B.~{Le Saux}, and S.~Lefèvre, ``Beyond rgb: Very high resolution
  urban remote sensing with multimodal deep networks,'' \emph{ISPRS Journal of
  Photogrammetry and Remote Sensing}, vol. 140, pp. 20--32, 2018, geospatial
  Computer Vision.

\bibitem{8789636}
P.~Shamsolmoali, M.~Zareapoor, R.~Wang, H.~Zhou, and J.~Yang, ``A novel deep
  structure u-net for sea-land segmentation in remote sensing images,''
  \emph{IEEE Journal of Selected Topics in Applied Earth Observations and
  Remote Sensing}, vol.~12, no.~9, pp. 3219--3232, 2019.

\bibitem{doi:10.1080/01431161.2020.1742944}
\BIBentryALTinterwordspacing
W.~Feng, H.~Sui, L.~Hua, C.~Xu, G.~Ma, and W.~Huang, ``Building extraction from
  vhr remote sensing imagery by combining an improved deep convolutional
  encoder-decoder architecture and historical land use vector map,''
  \emph{International Journal of Remote Sensing}, vol.~41, no.~17, pp.
  6595--6617, 2020. [Online]. Available:
  \url{https://doi.org/10.1080/01431161.2020.1742944}
\BIBentrySTDinterwordspacing

\bibitem{doi:10.1080/17538947.2020.1831087}
\BIBentryALTinterwordspacing
S.~Du, S.~Du, B.~Liu, and X.~Zhang, ``Incorporating deeplabv3+ and object-based
  image analysis for semantic segmentation of very high resolution remote
  sensing images,'' \emph{International Journal of Digital Earth}, vol.~14,
  no.~3, pp. 357--378, 2021. [Online]. Available:
  \url{https://doi.org/10.1080/17538947.2020.1831087}
\BIBentrySTDinterwordspacing

\bibitem{Strudel_2021_ICCV}
R.~Strudel, R.~Garcia, I.~Laptev, and C.~Schmid, ``Segmenter: Transformer for
  semantic segmentation,'' in \emph{Proceedings of the IEEE/CVF International
  Conference on Computer Vision (ICCV)}, October 2021, pp. 7262--7272.

\bibitem{10.1145/3505244}
S.~Khan, M.~Naseer, M.~Hayat, S.~W. Zamir, F.~S. Khan, and M.~Shah,
  ``Transformers in vision: A survey,'' \emph{ACM Comput. Surv.}, dec 2021,
  just Accepted.

\bibitem{https://doi.org/10.48550/arxiv.2002.09402}
\BIBentryALTinterwordspacing
A.~Fan, T.~Lavril, E.~Grave, A.~Joulin, and S.~Sukhbaatar, ``Addressing some
  limitations of transformers with feedback memory,'' 2020. [Online].
  Available: \url{https://arxiv.org/abs/2002.09402}
\BIBentrySTDinterwordspacing

\bibitem{pmlr-v97-tan19a}
M.~Tan and Q.~Le, ``{E}fficient{N}et: Rethinking model scaling for
  convolutional neural networks,'' in \emph{Proceedings of the 36th
  International Conference on Machine Learning}, ser. Proceedings of Machine
  Learning Research, K.~Chaudhuri and R.~Salakhutdinov, Eds., vol.~97.\hskip
  1em plus 0.5em minus 0.4em\relax PMLR, 09--15 Jun 2019, pp. 6105--6114.

\bibitem{https://doi.org/10.48550/arxiv.2010.11929}
\BIBentryALTinterwordspacing
A.~Dosovitskiy, L.~Beyer, A.~Kolesnikov, D.~Weissenborn, X.~Zhai,
  T.~Unterthiner, M.~Dehghani, M.~Minderer, G.~Heigold, S.~Gelly, J.~Uszkoreit,
  and N.~Houlsby, ``An image is worth 16x16 words: Transformers for image
  recognition at scale,'' 2020. [Online]. Available:
  \url{https://arxiv.org/abs/2010.11929}
\BIBentrySTDinterwordspacing

\bibitem{Sandler_2018_CVPR}
M.~Sandler, A.~Howard, M.~Zhu, A.~Zhmoginov, and L.-C. Chen, ``Mobilenetv2:
  Inverted residuals and linear bottlenecks,'' in \emph{Proceedings of the IEEE
  Conference on Computer Vision and Pattern Recognition (CVPR)}, June 2018.

\bibitem{doi:10.1080/10867651.2004.10487596}
\BIBentryALTinterwordspacing
A.~Telea, ``An image inpainting technique based on the fast marching method,''
  \emph{Journal of Graphics Tools}, vol.~9, no.~1, pp. 23--34, 2004. [Online].
  Available: \url{https://doi.org/10.1080/10867651.2004.10487596}
\BIBentrySTDinterwordspacing

\bibitem{labeling_2016}
``Isprs, 2016. international society for photogrammetry and remote sensing. 2d
  semantic labeling challenge,''
  \url{http://www2.isprs.org/commissions/comm3/wg4/semantic-labeling.html},
  accessed: 30-09-2021.

\bibitem{Gerke2014UseOT}
M.~Gerke, ``Use of the stair vision library within the isprs 2d semantic
  labeling benchmark (vaihingen),'' in \emph{University of Twente}, 01 2015,
  pp. 1,14.

\bibitem{Long_2015_CVPR}
J.~Long, E.~Shelhamer, and T.~Darrell, ``Fully convolutional networks for
  semantic segmentation,'' in \emph{Proceedings of the IEEE Conference on
  Computer Vision and Pattern Recognition (CVPR)}, June 2015.

\bibitem{8309343}
Z.~Zhang, Q.~Liu, and Y.~Wang, ``Road extraction by deep residual u-net,''
  \emph{IEEE Geoscience and Remote Sensing Letters}, vol.~15, no.~5, pp.
  749--753, 2018.

\bibitem{7803544}
V.~Badrinarayanan, A.~Kendall, and R.~Cipolla, ``Segnet: A deep convolutional
  encoder-decoder architecture for image segmentation,'' \emph{IEEE
  Transactions on Pattern Analysis and Machine Intelligence}, vol.~39, no.~12,
  pp. 2481--2495, 2017.

\bibitem{8100143}
H.~Zhao, J.~Shi, X.~Qi, X.~Wang, and J.~Jia, ``Pyramid scene parsing network,''
  in \emph{2017 IEEE Conference on Computer Vision and Pattern Recognition
  (CVPR)}, 2017, pp. 6230--6239.

\bibitem{9491802}
H.~Chen, Z.~Qi, and Z.~Shi, ``Remote sensing image change detection with
  transformers,'' \emph{IEEE Transactions on Geoscience and Remote Sensing},
  vol.~60, pp. 1--14, 2022.

\bibitem{Peng_2017_CVPR}
C.~Peng, X.~Zhang, G.~Yu, G.~Luo, and J.~Sun, ``Large kernel matters -- improve
  semantic segmentation by global convolutional network,'' in \emph{Proceedings
  of the IEEE Conference on Computer Vision and Pattern Recognition (CVPR)},
  July 2017.

\bibitem{7301381}
S.~Paisitkriangkrai, J.~Sherrah, P.~Janney, and A.~Van-Den~Hengel, ``Effective
  semantic pixel labelling with convolutional networks and conditional random
  fields,'' in \emph{2015 IEEE Conference on Computer Vision and Pattern
  Recognition Workshops (CVPRW)}, 2015, pp. 36--43.

\bibitem{https://doi.org/10.48550/arxiv.1606.02585}
\BIBentryALTinterwordspacing
J.~Sherrah, ``Fully convolutional networks for dense semantic labelling of
  high-resolution aerial imagery,'' 2016. [Online]. Available:
  \url{https://arxiv.org/abs/1606.02585}
\BIBentrySTDinterwordspacing

\bibitem{rs9050446}
\BIBentryALTinterwordspacing
H.~Wang, Y.~Wang, Q.~Zhang, S.~Xiang, and C.~Pan, ``Gated convolutional neural
  network for semantic segmentation in high-resolution images,'' \emph{Remote
  Sensing}, vol.~9, no.~5, 2017. [Online]. Available:
  \url{https://www.mdpi.com/2072-4292/9/5/446}
\BIBentrySTDinterwordspacing

\bibitem{10.1007/978-3-642-75988-8_28}
M.~Holschneider, R.~Kronland-Martinet, J.~Morlet, and P.~Tchamitchian, ``A
  real-time algorithm for signal analysis with the help of the wavelet
  transform,'' in \emph{Wavelets}, J.-M. Combes, A.~Grossmann, and
  P.~Tchamitchian, Eds.\hskip 1em plus 0.5em minus 0.4em\relax Berlin,
  Heidelberg: Springer Berlin Heidelberg, 1990, pp. 286--297.

\bibitem{9102424}
L.~Ding, H.~Tang, and L.~Bruzzone, ``Lanet: Local attention embedding to
  improve the semantic segmentation of remote sensing images,'' \emph{IEEE
  Transactions on Geoscience and Remote Sensing}, vol.~59, no.~1, pp. 426--435,
  2021.

\bibitem{9686732}
C.~Zhang, W.~Jiang, Y.~Zhang, W.~Wang, Q.~Zhao, and C.~Wang, ``Transformer and
  cnn hybrid deep neural network for semantic segmentation of
  very-high-resolution remote sensing imagery,'' \emph{IEEE Transactions on
  Geoscience and Remote Sensing}, vol.~60, pp. 1--20, 2022.

\bibitem{doi:10.1080/2150704X.2021.1910362}
\BIBentryALTinterwordspacing
K.~Yan, H.~Wang, S.~Bu, L.~Yang, and J.~Li, ``Scene parsing for very high
  resolution remote sensing images using on attention-residual block-embedded
  adversarial networks,'' \emph{Remote Sensing Letters}, vol.~12, no.~7, pp.
  625--635, 2021. [Online]. Available:
  \url{https://doi.org/10.1080/2150704X.2021.1910362}
\BIBentrySTDinterwordspacing

\bibitem{doi:10.1080/01431161.2021.2018147}
\BIBentryALTinterwordspacing
S.~Zhang, C.~Wang, J.~Li, and Y.~Sui, ``Mf-dfnet: a deep learning method for
  pixel-wise classification of very high-resolution remote sensing images,''
  \emph{International Journal of Remote Sensing}, vol.~43, no.~1, pp. 330--348,
  2022. [Online]. Available:
  \url{https://doi.org/10.1080/01431161.2021.2018147}
\BIBentrySTDinterwordspacing

\bibitem{9690154}
Y.~Su, J.~Cheng, W.~Wang, H.~Bai, and H.~Liu, ``Semantic segmentation for
  high-resolution remote-sensing images via dynamic graph context reasoning,''
  \emph{IEEE Geoscience and Remote Sensing Letters}, vol.~19, pp. 1--5, 2022.

\bibitem{9681903}
L.~Wang, R.~Li, C.~Duan, C.~Zhang, X.~Meng, and S.~Fang, ``A novel transformer
  based semantic segmentation scheme for fine-resolution remote sensing
  images,'' \emph{IEEE Geoscience and Remote Sensing Letters}, vol.~19, pp.
  1--5, 2022.

\bibitem{9538389}
W.~Zhou, J.~Jin, J.~Lei, and J.-N. Hwang, ``Cegfnet: Common extraction and gate
  fusion network for scene parsing of remote sensing images,'' \emph{IEEE
  Transactions on Geoscience and Remote Sensing}, vol.~60, pp. 1--10, 2022.

\bibitem{9519842}
X.~Zheng, X.~Wu, L.~Huan, W.~He, and H.~Zhang, ``A gather-to-guide network for
  remote sensing semantic segmentation of rgb and auxiliary image,'' \emph{IEEE
  Transactions on Geoscience and Remote Sensing}, vol.~60, pp. 1--15, 2022.

\bibitem{9258402}
R.~Liu, L.~Mi, and Z.~Chen, ``Afnet: Adaptive fusion network for remote sensing
  image semantic segmentation,'' \emph{IEEE Transactions on Geoscience and
  Remote Sensing}, vol.~59, no.~9, pp. 7871--7886, 2021.

\bibitem{9345482}
A.~Li, L.~Jiao, H.~Zhu, L.~Li, and F.~Liu, ``Multitask semantic boundary
  awareness network for remote sensing image segmentation,'' \emph{IEEE
  Transactions on Geoscience and Remote Sensing}, vol.~60, pp. 1--14, 2022.

\bibitem{rs10091429}
\BIBentryALTinterwordspacing
S.~Piramanayagam, E.~Saber, W.~Schwartzkopf, and F.~W. Koehler, ``Supervised
  classification of multisensor remotely sensed images using a deep learning
  framework,'' \emph{Remote Sensing}, vol.~10, no.~9, 2018. [Online].
  Available: \url{https://www.mdpi.com/2072-4292/10/9/1429}
\BIBentrySTDinterwordspacing

\bibitem{10.1117/12.2243169}
\BIBentryALTinterwordspacing
S.~Piramanayagam, W.~Schwartzkopf, F.~W. Koehler, and E.~Saber,
  ``{Classification of remote sensed images using random forests and deep
  learning framework},'' in \emph{Image and Signal Processing for Remote
  Sensing XXII}, L.~Bruzzone and F.~Bovolo, Eds., vol. 10004, International
  Society for Optics and Photonics.\hskip 1em plus 0.5em minus 0.4em\relax
  SPIE, 2016, pp. 205 -- 212. [Online]. Available:
  \url{https://doi.org/10.1117/12.2243169}
\BIBentrySTDinterwordspacing

\bibitem{8727958}
K.~Nogueira, M.~Dalla~Mura, J.~Chanussot, W.~R. Schwartz, and J.~A. dos Santos,
  ``Dynamic multicontext segmentation of remote sensing images based on
  convolutional networks,'' \emph{IEEE Transactions on Geoscience and Remote
  Sensing}, vol.~57, no.~10, pp. 7503--7520, 2019.

\end{thebibliography}
%



%





\end{document}